\documentclass[runningheads]{llncs}

\usepackage{eccv}

\usepackage{eccvabbrv}

\usepackage{graphicx}
\usepackage{booktabs}
\usepackage{multirow}
\usepackage{amssymb}

\usepackage[accsupp]{axessibility}  

\usepackage{hyperref}

\usepackage{orcidlink}
\usepackage{subcaption}
\usepackage{pifont}
\usepackage{amsmath}
\newcommand{\cmark}{\ding{51}}%

\begin{document}

\title{Chronologically Accurate Retrieval for \\ Temporal Grounding of Motion-Language Models} 

\titlerunning{Chronologically Accurate Retrieval for Motion-Language Models}

\author{Kent Fujiwara\orcidlink{0000-0002-2205-6115} \and
Mikihiro Tanaka\orcidlink{0000-0002-8259-6658} \and
Qing Yu\orcidlink{0000-0001-6965-9581}}

\authorrunning{K.~Fujiwara et al.}

\institute{LY Corporation, Tokyo, Japan \\
\email{\{kent.fujiwara, mikihiro.tanaka, yu.qing\}@lycorp.co.jp}}

\maketitle

\begin{abstract}
With the release of large-scale motion datasets with textual annotations, the task of establishing a robust latent space for language and 3D human motion has recently witnessed a surge of interest. Methods have been proposed to convert human motion and texts into features to achieve accurate correspondence between them. Despite these efforts to align language and motion representations, we claim that the temporal element is often overlooked, especially for compound actions, resulting in chronological inaccuracies. To shed light on the temporal alignment in motion-language latent spaces, we propose Chronologically Accurate Retrieval (CAR) to evaluate the chronological understanding of the models. We decompose textual descriptions into events, and prepare negative text samples by shuffling the order of events in compound action descriptions. We then design a simple task for motion-language models to retrieve the more likely text from the ground truth and its chronologically shuffled version. CAR reveals many cases where current motion-language models fail to distinguish the event chronology of human motion, despite their impressive performance in terms of conventional evaluation metrics. To achieve better temporal alignment between text and motion, we further propose to use these texts with shuffled sequence of events as negative samples during training to reinforce the motion-language models. We conduct experiments on text-motion retrieval and text-to-motion generation using the reinforced motion-language models, which demonstrate improved performance over conventional approaches, indicating the necessity to consider temporal elements in motion-language alignment.
\keywords{Human motion analysis \and Motion-language model \and Text-motion retrieval \and Text-motion generation \and Motion chronology}
\end{abstract} 
\section{Introduction}
\label{sec:intro}

Modeling human motion has become an important task in computer vision, especially with the emergence of applications for animating gaming characters and online avatars. To deal with the complex nature of human motion, natural language is gaining popularity as a medium to describe skeletal human motion, leading to 3D human skeletal motion datasets annotated with textual descriptions. This has led to proposals of novel tasks that attempt to establish a joint latent space for motion and language, including motion-text retrieval~\cite{petrovich2023tmr, tevet2022motionclip,yu2024exploring}, and motion generation from texts~\cite{athanasiou2022teach, chen2023executing, ghosh2021synthesis, kalakonda2022action, petrovich2022temos, tevet2022human, zhang2023t2m, zhang2022motiondiffuse}.

\begin{figure}
\vspace{-5pt}
    \centering
    \includegraphics[width=\linewidth]{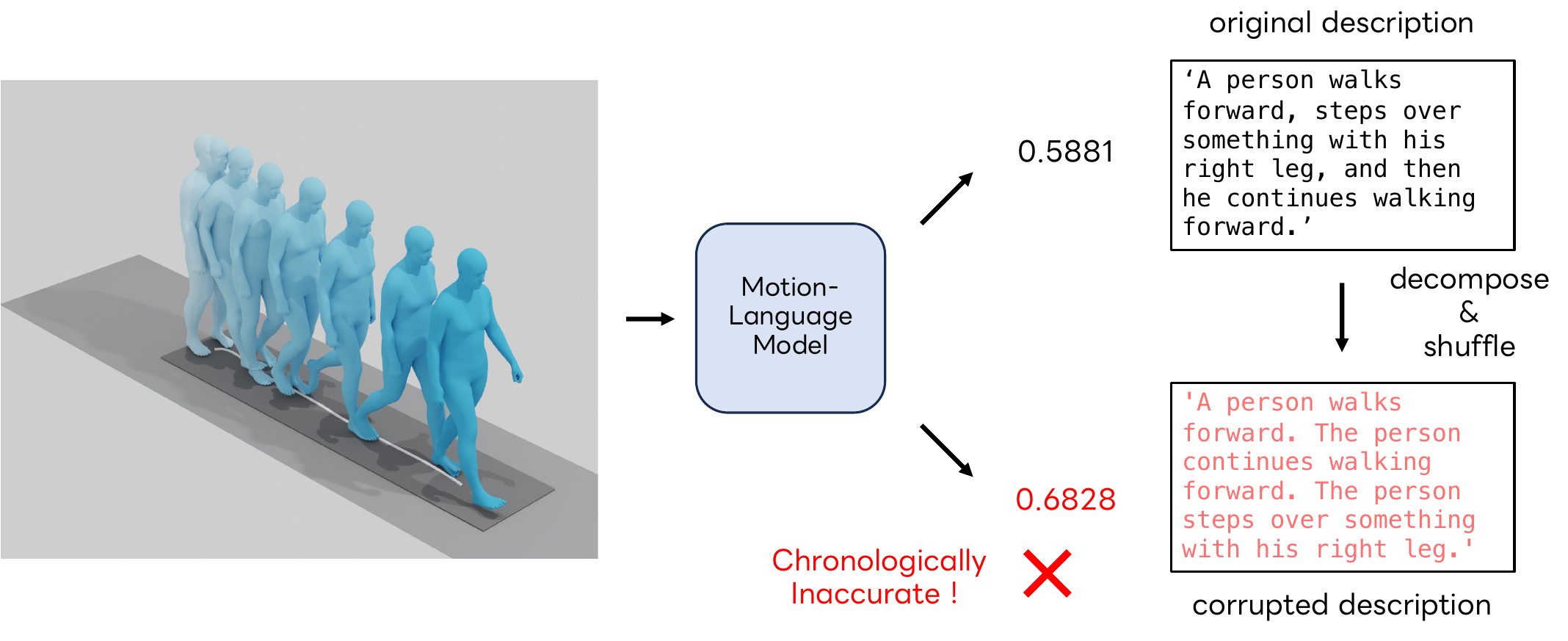}
    \caption{Overview of Chronologically Accurate Retrieval test. Given a motion sequence, motion-language models trained on text-motion datasets are asked to retrieve the more relevant text from the ground truth and its shuffled version. Original texts are decomposed into events by off-the-shelf Large Language Models, which are randomly shuffled.}
    \label{fig:CAR}
    \vspace{-5pt}
\end{figure}

Although various methods~\cite{petrovich2023tmr, tevet2022motionclip, yu2024exploring} demonstrate impressive performance in this field, we claim that the temporal element is not fully considered in most of the motion-language models. Establishing temporal alignment between motion and text becomes important when there are multiple events in a motion sequence, especially in motion recognition and generation, where the order of events needs to be accurate. 
However, there is currently no method that explicitly evaluates whether temporal alignment between motion and text is achieved or not. 

To reveal the performance of the current state-of-the-art motion-language model in terms of understanding the chronology of events, we propose a simple test, which we call Chronologically Accurate Retrieval (CAR). This test assesses whether the motion-language model is capable of recognizing the order of events. We decompose the original motion descriptions into events, which are then shuffled to create chronologically inaccurate descriptions. We test the model to see whether it can retrieve the chronologically accurate description. The overview of CAR is shown in Fig.~\ref{fig:CAR}. CAR reveals that the current model falls short of understanding the temporal component of motion and descriptions. We also combine the state-of-the-art model with various language models to observe the tendencies among language models when used in motion-language models.

To address the issue of temporal alignment between motion and text, we further propose a simple strategy to refine the motion-language model. We enhance the current contrastive learning framework with the chronologically negative text samples. We employ the description derived from shuffled events as an incorrect description of the original motion, and append these shuffled descriptions as negative text samples. Using these negative samples, the motion-language model is trained to differentiate chronologically accurate descriptions from the incorrect ones, to achieve stronger temporal alignment between the two modalities.

We evaluate the refined motion-language model through text-motion retrieval and motion generation from texts. The results reveal that our proposal enhances the performance of the motion-language model in both of these tasks.

The contributions of our research are as follows:

\begin{itemize}
    \item We propose to evaluate the temporal alignment between motion and language through Chronologically Accurate Retrieval (CAR), a novel test that measures how accurate models can differentiate original motion descriptions from chronologically incorrect ones given a motion.
    \item We reveal that current motion-language models fail to fully comprehend the temporal component of motion and language through CAR, even when larger language models are introduced to the motion-language model.
    \item We propose a simple solution to achieve better temporal correspondence between language and motion, where shuffled event descriptions are used as negative text samples to train and refine a motion-language model through contrastive learning. The resulting motion-language model achieves high performance in both text-motion retrieval and motion generation from text.
\end{itemize}

\section{Related Work}
\label{sec:related}

\subsection{Motion-Language Models}

With the recent progress in cross-modal representation and the success of recent language models, language now plays a significant role in computer vision tasks. A vast amount of image data annotated with textual descriptions is employed in tasks such as image captioning~\cite{li2022blip, li2023blip}, segmentation~\cite{kirillov2023segment}, and recognition~\cite{radford2021learning}. The development of diffusion models~\cite{dhariwal2021diffusion, rombach2022high} has also led to an explosion of interest in attempting to generate images from textual descriptions~\cite{rombach2022high}.   

The success of the cross-modal analysis in the image domain has recently garnered interest in establishing a cross-modal model connecting language and 3D human motion. Earlier research used 3D human motion datasets solely for computer vision tasks, such as action recognition~\cite{shahroudy2016ntu}, pose estimation~\cite{ionescu2013human3}, and motion prediction~\cite{liu2022pisep}. However, there has also been some interest in trying to annotate human motion with textual descriptions, as text is a convenient medium to describe the complex and varying nature of human motion. KIT-ML~\cite{plappert2016kit} is one of the first datasets to add textual description to 3D human motion data captured using motion capture devices. It is a collection of human motions consisting mainly of locomotion-related actions. Recently, HumanML3D~\cite{guo2022generating} was released with textual annotations added to a large collection of motion data consisting of AMASS~\cite{mahmood2019amass} and HumanAct12~\cite{guo2020action2motion} datasets. There are other small-scale motion datasets focusing on specific tasks, such as hand-object manipulation~\cite{taheri2020grab}. There are also attempts to enhance these datasets by capturing 3D human motion from videos and annotating the descriptions~\cite{lin2023motionx}.

With the release of these larger-scale, motion-language datasets, there are various ongoing research efforts that attempt to capitalize on such datasets, as well as the capability of recent large language models (LLMs)~\cite{brown2020language, touvron2023llama}. As skeletal human motion data focuses solely on human motion without any background information, motion recognition~\cite{yan2018spatial, shi2019two, duan2022revisiting} is one of the most popular research topics. There are tasks that attempt to classify what the motion is representing, which has recently evolved into retrieval tasks~\cite{petrovich2023tmr}, where given a text or motion, the task is to find the most appropriate counterpart. 
Similar to the image domain, generating novel motion from semantic description~\cite{petrovich2022temos, tevet2022human, zhang2023t2m, zhang2022motiondiffuse} has also become a popular task. Beginning from action labels~\cite{petrovich2021action}, the models have evolved to generate appropriate motions corresponding to given textual descriptions.

All of these tasks require establishing a robust motion-language latent space, where similar descriptions are located close to its corresponding motion representations. Despite the recent advances, there has been relatively little interest in exploring the temporal aspect of motions. Some methods~\cite{miki2020weakly, yu2023frame} attempt to localize the weakly assigned action labels to segments within the motion sequence to discover what actions are taking place where. Other attempts~\cite{Shafir2024PriorMDM,Petrovich2024stmc} have proposed to generate compound motions by connecting different actions smoothly. However, this does not lead to a common motion-language latent space, which is capable of representing the temporal component between language and motion. In this paper, we focus on the temporal alignment between text and motion, and demonstrate how the current motion-language models either misinterpret or overlook the temporal component of motion and language.

\subsection{Alignment of Cross-Modality Latent Space}

As previously stated, it is crucial for cross-modal models to establish accurate relationships between different representations. Achieving alignment between language and other modalities has become one of the major focal points of ongoing research. For example, in the image domain, some research has attempted to align text with images~\cite{chefer2023attend} so that accurate correspondence is achieved between the given caption and the generated image. These models are used for various downstream tasks such as segmentation, as the attention corresponding to the text input in the generation process can be interpreted as the region of interest~\cite{burgert2022peekaboo}. Some studies also tried to discover misalignment between the modalities. One work of research~\cite{yuksekgonul2022and} identified flaws in the popular CLIP model~\cite{radford2021learning}, where shuffling subject and object or other various elements in the texts did not lead to a substantial decline in the performance of the model.

There are also attempts to align text with sequential data. For example, in the field of video analysis, where certain events are localized within videos, some methods~\cite{ko2022video, zhang2023exploring, jung2023overcoming} attempt to identify the proper chronological correspondence between textual descriptions of events and videos. In the field of audio data analysis, there are also attempts to use language models to provide accurate descriptions to generate the corresponding sound sequences~\cite{borsos2023audiolm, kreuk2022audiogen}. In the field of human motion, as there is a lack of data that has frame-wise annotations indicating what actions are taking place at each moment, the temporal component of motions is currently being overlooked. Inspired by the recent research in these related fields, we uncover the deficiencies of the current motion-language models, and propose a simple solution to alleviating the temporal alignment issue.

\section{Temporal Element in Motion-Language Latent Space}
\label{sec:analysis}

We first focus on analyzing the temporal element in motion-language latent space established by the state-of-the-art motion-language model. We choose the task of text-to-motion/motion-to-text retrieval to assess the alignment of the two modalities. We employ TMR~\cite{petrovich2023tmr}, which is a recently proposed text-motion retrieval model that uses contrastive learning and a generative model to establish the correspondence between features of texts and human motion sequences.

\subsection{Text-Motion Retrieval}
Given a text description $T$, the goal of text-to-motion retrieval is to find a 3D skeletal motion sequence $M$ that most closely resembles the provided text. Motion-to-text retrieval conducts the opposite task, where the closest textual description is retrieved when given a motion sequence. 

TMR proposes to achieve this by following CLIP~\cite{radford2021learning}, training a model to learn a function $f(z^T,z^M)$ that provides the similarity between the outputs of motion and textual encoders. Here, $z^T$ is a language feature obtained from a textual encoder, and $z^M$ is the motion feature from a motion encoder. To train this model, given $N$ samples of text-motion feature pairs $(z^T_1,z^M_1) \dots (z^T_N,z^M_N)$, the method calculates cosine similarities between all the combinations, $S_{ij}=(z^T_i,z^M_j)$, to establish a similarity matrix of features $\mathbf{S} \in \mathbb{R}^{N\times N}$. The similarity matrix is used to optimize the contrastive loss:
\begin{equation}
\mathcal{L}=-\frac{1}{2 N} \sum_i\left(\log \frac{\exp S_{i i} / \tau}{\sum_j \exp S_{i j} / \tau}+\log \frac{\exp S_{i i} / \tau}{\sum_j \exp S_{j i} / \tau}\right)\,,
\label{eq:contrast}
\end{equation}
where $\tau$ is a temperature parameter. With the support of additional constraints, the model learns to bring together motions and texts that correspond to each other, while returning low similarity values to the wrong pairs. 

\subsection{Chronologically Accurate Retrieval}

TMR demonstrates robust performance in text-motion retrieval tasks. However, the analysis does not reveal whether the model is capable of understanding the temporal relationships between language and motion. This is due to the scale of the dataset, where the variety in the motion sequences does not necessarily consider chronological variations. Although motion datasets are usually augmented by mirroring the human body pose, motion augmentation by changing the order of actions is a challenging task, likely resulting in undesirable noise.

\begin{figure}[t]
    \begin{minipage}[b]{0.4\linewidth}
        \centering
        \includegraphics[keepaspectratio, scale=0.25]{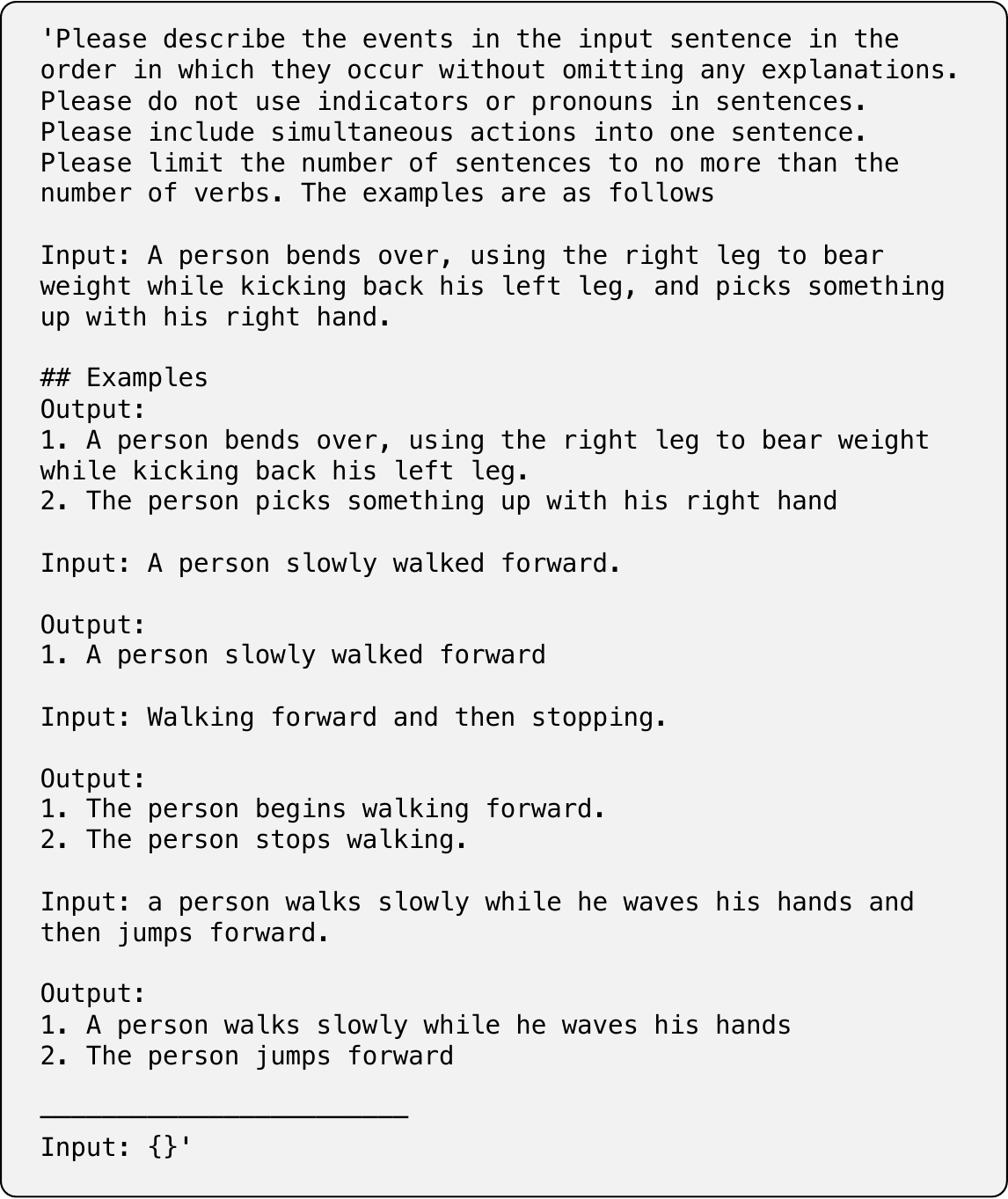}
        \subcaption{}\label{fig:GPTcom}
        \end{minipage}
    \begin{minipage}[b]{0.4\linewidth}
        \centering
        \includegraphics[keepaspectratio, scale=0.22]{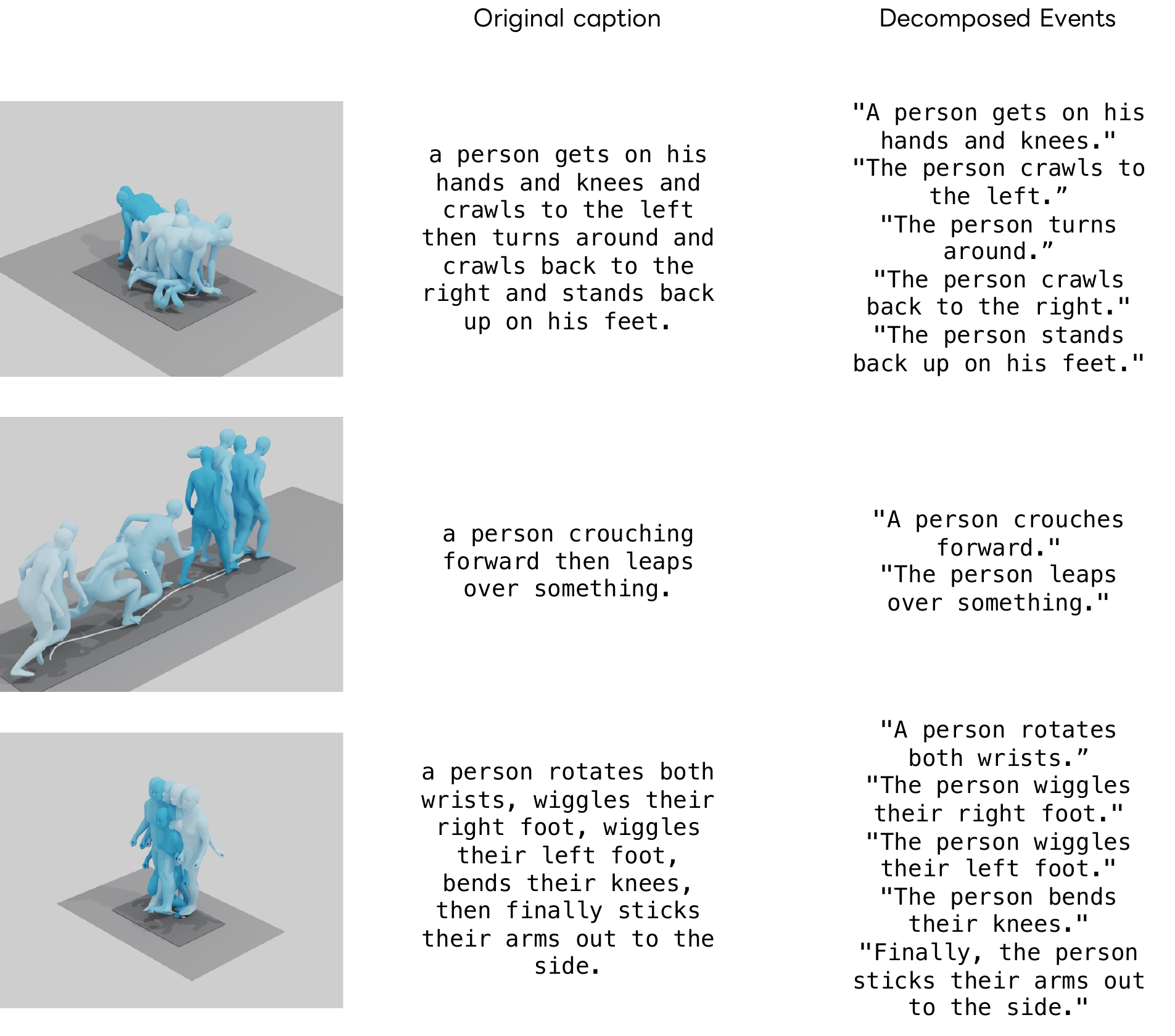}
        \subcaption{}\label{fig:GPTeg}
    \end{minipage}
    
\caption{(a) Command prompt used to decompose motion captions into events. We provide the prompt as input to a Large Language Model, GPT3.5, and add the captions in the dataset to decompose descriptions into events. (b) Examples of original motion captions and decomposed events. Corresponding motions are shown on the left.}
\label{fig:GPT}
\vspace{-5pt}
\end{figure}

To analyze the chronological understanding of motion-language models, we focus on language and propose a simple task that tests whether the models understand the temporal element of the motion-language relationship. For this task, we propose to decompose the original textual descriptions provided for the motions into events contained in the description. We conduct the event decomposition using an off-the-shelf LLM, GPT3.5. The command prompt to generate the decomposed events is shown in Fig.~\ref{fig:GPTcom}. By presenting exemplar decompositions in the prompt, the LLM is able to decompose textual descriptions of motions into sequences of events, as presented in Fig.~\ref{fig:GPTeg}.

We then form a pair of descriptions, consisting of the ground truth text and a chronologically incorrect description. The incorrect text is obtained by shuffling the events, if more than one, to generate a new description with a different chronological order. We then provide the motion-language model with a motion and calculate the similarity between the motion feature and the two descriptions, and choose the one with the higher likelihood to be the likely description for the motion. We call this task, Chronologically Accurate Retrieval (CAR) test.

\subsection{Analysis}

We conduct the Chronologically Accurate Retrieval test on the HumanML3D dataset~\cite{guo2022generating}. The dataset consists of motions from AMASS~\cite{mahmood2019amass} and HumanAct12~\cite{guo2020action2motion} motion datasets, and textual descriptions assigned to each of the motion sequences. Following prior works, we also augment the dataset by mirroring the human body pose and assigning them with the adjusted descriptions. The official dataset split consists of 23,384, 1,460, and 4,380 motion sequences for training, validation, and test sets, respectively. As each motion contains 3.0 descriptions on average, we select one random description during training and the first one for testing. For each description corresponding to each motion sequence, we apply the decomposition using LLM to obtain the event descriptions. Out of 4,380 test set motions, multiple events are found in 2,677 sequences. In this section, we only use this test subset of multiple event sequences to conduct CAR. For this test, we prepare the chronologically inaccurate sample for each sequence in the test subset by shuffling the order of the events, and concatenating them as a single text input. Given the original texts $\{T_1, \dots T_K\}$ and the chronologically inaccurate samples $\{C_1, \dots C_K\}$, CAR is calculated as 
\begin{equation}
CAR=\frac{1}{K} \sum_i^K g\left(f(z^T_i,z^M_i),f(z^C_i,z^M_i)\right)\,,
\label{eq:car}
\end{equation}
where $z^C$ is the feature of the chronologically inaccurate text, and the function $g$ returns $1$ if $f(z^T_i,z^M_i)$ is larger, and $0$ otherwise.

As decomposed events could have different structural characteristics from the original texts, which may lead to an undesirable domain gap, we prepare two scenarios. The first, ``orig $\rightarrow$ event'', is the case where original HumanML3D captions are used as is for input. 
During the test, the model is tasked with distinguishing the original texts in the test set from the shuffled event descriptions obtained from decomposing and shuffling the same text. The second, ``event $\rightarrow$ event'', uses decomposed event descriptions concatenated into one text in the correct order to train the model. 
For testing, the model is asked to compare the correctly concatenated events, and its shuffled version.  

As language plays an important component in this test, we also enhance TMR by employing different textual encoders to see whether textual embedding has an effect on the resulting motion-language latent space. We compare DistilBERT~\cite{sanh2019distilbert}, the default language model used in TMR, with CLIP~\cite{radford2021learning}, which is a popular choice in motion analysis. Additionally, we introduce two variants of T5 language models~\cite{raffel2020exploring}, t5-base and t5-large, to observe whether the size of models has an effect. The token embeddings from the language models are concatenated with the learnable parameters to be used in the VAE-style encoder.

We mostly follow the settings of the original TMR. For motion representation, we use the representation consisting of relative joint positions and accelerations employed by Guo et al.~\cite{guo2020action2motion}, as in the original TMR. We maintain the same network structure as the original method, using two VAE-based encoders to derive features for motion sequences and texts. A motion decoder is also used to decode the original motion from both the features of the corresponding motion and text. The weights balancing the losses to train TMR are also kept the same as the original. We set the batch size to 32 in order to achieve better results for all the patterns using different language models under the same environment, and employ the AdamW optimizer~\cite{loshchilov2018decoupled} with the learning rate set at $0.0001$. The setting led to more accurate results for the original method as well. Further details can be found in the supplementary material.

We show the performance in terms of CAR in Table~\ref{tab:TMRnegret}.  The first 4 rows of the rightmost column of each scenario show CAR for TMR equipped with different language models. Despite its success in the retrieval of the HumanML3D dataset, the model struggles to differentiate the original text from decomposed and shuffled event descriptions, as most models achieve success around $60\%$ of the time, only marginally above the chance level of $50\%$. CAR accuracy does not show much change in ``event $\rightarrow$ event'' scenario either. This indicates that the current motion-language models fail to capture the intricate temporal relationship between language and motion. This issue of temporal alignment must be addressed to further enhance the performance of future motion-language models.

\section{Contrastive Learning with Shuffled Events}

In this paper, we take the initiative and propose a simple solution to achieving better correspondence between textual descriptions and motions in terms of chronology. Inspired by the recent research that attempts to establish better correspondence between language and images in contrastive learning~\cite{yuksekgonul2022and}, we propose to reinforce the model by incorporating chronologically incorrect descriptions obtained by shuffling the decomposed events in the previous analysis.

\begin{figure*}[t]
\vspace{-5pt}
    \centering
    \includegraphics[width=\linewidth]{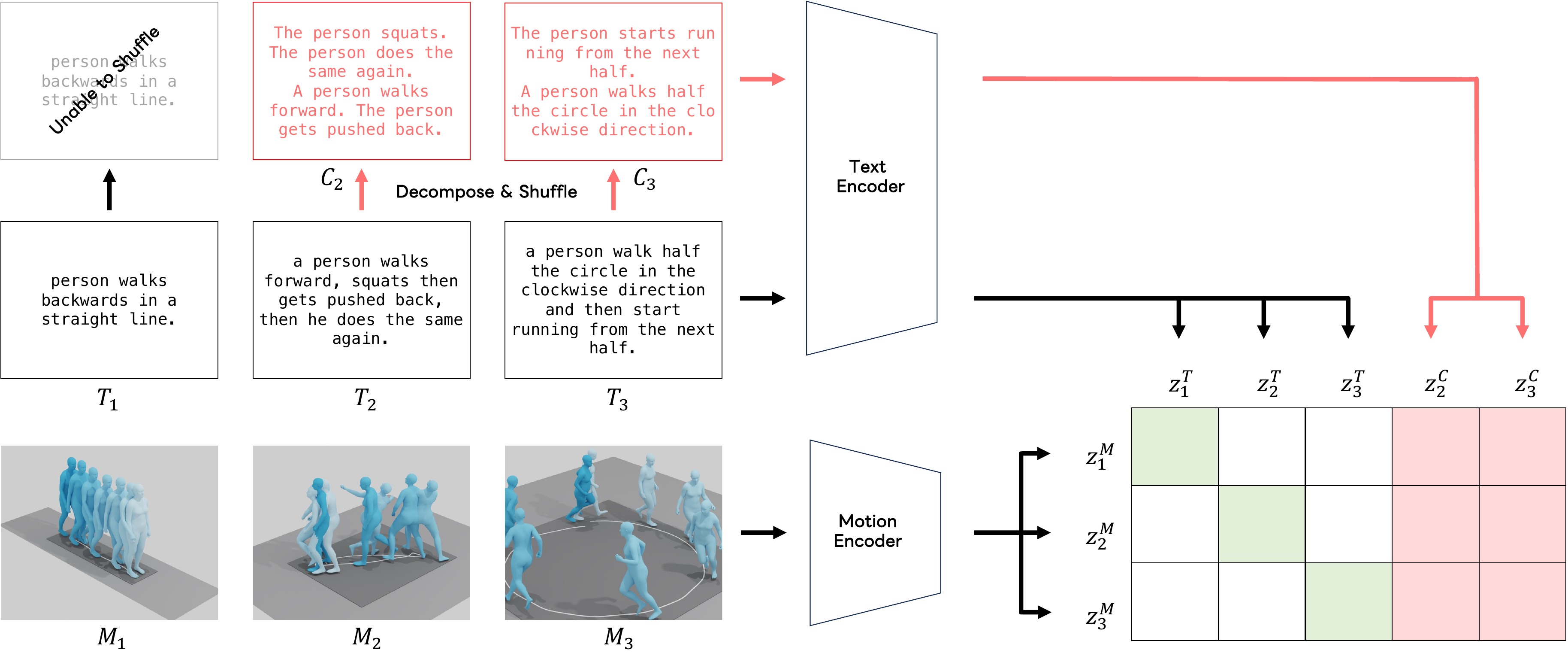}
    \caption{Overview of the proposed contrastive learning scheme with chronological negative samples. We use the texts derived from shuffling the event order and employ them as negative text samples, corresponding to items indicated in pink. }
    \label{fig:Negative}
    \vspace{-5pt}
\end{figure*}

\subsection{Shuffled Events as Hard Negative Chronological Samples}
Because the event-based shuffled descriptions are accurate in terms of the words used but not in terms of chronological order, they serve as crucial resource for motion-language models to recognize the inaccuracies in the chronology of events. We, therefore, propose to utilize these shuffled event descriptions as the hard negative samples corresponding to the original motion.

We consider a batch of $N$ texts $\{T_1, \dots T_N\}$ and motions $\{M_1, \dots M_N\}$. Before encoding them into features and calculating the similarities between them, we select texts that contain more than one event and decompose them into event descriptions following the method described in Section~\ref{sec:analysis}. Here, we indicate the number of text-motion pairs that contain multiple events as $K$. The decomposed $K$ texts are then shuffled randomly to obtain chronologically incorrect descriptions $C_i$, which corresponds to the original text $T_i$.

Once the shuffled texts $\{C_1, \dots C_K\}$ are obtained, we calculate the similarity matrix in the same manner as TMR and other CLIP-based contrastive learning approaches. However, instead of the similarity matrix being $\mathbf{S} \in \mathbb{R}^{N\times N}$, our modified similarity matrix is $\tilde{\mathbf{S}} \in \mathbb{R}^{N\times (N+K)}$, with additional columns corresponding to the shuffled text features $z^C_i$. 
We calculate the same row-wise and column-wise cross entropy loss, but exclude the column-wise loss for columns corresponding to the shuffled texts, as they do not have any corresponding motion. Therefore, the loss Eq.~(\ref{eq:contrast}) becomes
\begin{equation}
\mathcal{L}= \mathcal{L}_{t2m} + \mathcal{L}_{m2t}\,,
\end{equation}
where

\begin{equation}
     \mathcal{L}_{t2m} = -\frac{1}{N} \sum_i^N \log \frac{\exp \tilde{S}_{i i} / \tau}{\sum_j^{N} \exp \tilde{S}_{i j} / \tau}\,,
\end{equation}
\begin{equation}
     \mathcal{L}_{m2t} = -\frac{1}{N} \sum_i^{N} \log \frac{\exp \tilde{S}_{i i} / \tau}{\sum_j^{(N+K)} \exp \tilde{S}_{j i} / \tau},
\end{equation}
and $\tilde{S}_{i,j}$ is the $i,j$-the element in the similarity matrix $\tilde{\mathbf{S}}$.
This allows the model to use the shuffled events as negative samples, and train representations that differentiate chronologically corrupted descriptions from the true ones. An overview of the proposed scheme using shuffled texts as negative samples is shown in Fig.~\ref{fig:Negative}

\begin{table}[t]
\caption{Comparison of CAR accuracy and motion-to-text retrieval results with both the original and corrupted texts. We insert chronologically inaccurate texts as candidates for retrieval. Ours indicate models trained with the hard negative samples.}
\label{tab:TMRnegret}
    \centering
    \setlength{\tabcolsep}{4pt}
    \resizebox{0.99\linewidth}{!}{
    \begin{tabular}{llc|cccc|c|cccc|c}
    \toprule & & & \multicolumn{10}{c}{Motion-to-text retrieval}\\ 
    \multirow{2}{*}{Method} & \multirow{2}{*}{ Encoder}& Train with  &  \multicolumn{5}{c}{  ``orig $\rightarrow$ event''}  &\multicolumn{5}{|c}{ ``event $\rightarrow$ event''}\\
      & & Negatives &  $\mathrm{R} @ 1 \uparrow$ &  R@5$\uparrow$ & $\mathrm{R} @ 10 \uparrow$ & MedR $\downarrow$&CAR &$\mathrm{R} @ 1 \uparrow$ & R@5$\uparrow$ & $\mathrm{R} @ 10 \uparrow$ & MedR $\downarrow$& CAR\\ 
    \hline\hline 
     \multirow{4}{*}{ TMR~\cite{petrovich2023tmr}} &DistilBERT&  & 7.89 & 19.82 & 28.40 & 33.75 & 64.81& 6.93 & 17.08 & 26.23 & 36.00 & 65.04 \\
    &CLIP & &  6.18 & 16.54 & 24.48 & 43.00 & 63.17 &6.73 & 17.63 & 26.89 & 40.25&62.87\\
    &t5-base&  &6.71 & 17.63 & 28.15 & 35.00 & 66.42 & 6.14 & 15.58 & 24.02 & 36.00 & 63.91\\
    &t5-large& &7.76 & 20.12 & 29.36 & 34.00 & 66.72 &5.86 & 16.40 & 25.84 & 35.00 & 66.83 \\\hline
     \multirow{4}{*}{ Ours } &DistilBERT&\cmark &9.38 &  \textbf{23.31} & \textbf{34.10} & \textbf{24.00}&99.33& \textbf{8.90} &  \textbf{21.49} & \textbf{31.68} & \textbf{27.50} &92.90\\
     & CLIP&\cmark & 8.19 & 20.69 & 30.36 & 29.50 & 98.88& 8.01 & 20.28 & 29.65 & 34.00 & 91.37 \\
     &t5-base&\cmark  & 8.55 & 23.08 & 33.10 & 25.00& \textbf{99.74}& 7.6 & 20.14 & 30.13 & 30.00 & 91.78\\
     &t5-large &\cmark & \textbf{9.65}  & 22.88 & 32.71 & 25.00 & \textbf{99.74} & 6.91 & 19.89 & 29.29 & \textbf{27.50} & \textbf{93.09}\\
    
    \hline

    \end{tabular}}
    \vspace{-5pt}
\end{table}

\subsection{Text-Motion Retrieval}

To evaluate the effectiveness of our approach, we conduct the text-motion retrieval task by including the chronologically inaccurate texts as hard negative samples. We follow the same protocol as the previous experiment. We employ the same variations of TMR and train them using the additional negative samples.  

\textbf{CAR Accuracy.} Firstly, we show whether the proposed training scheme achieves robustness in terms of chronological accuracy. The bottom half of the rightmost column of each scenario in Table~\ref{tab:TMRnegret} shows the CAR accuracy of the motion-language model trained by our proposed method. In both ``orig $\rightarrow$ event'' and ``event $\rightarrow$ event'', we employ the shuffled event descriptions as hard negative samples to train the model in the aforementioned manner. Remarkably, all the models successfully learned to distinguish original texts from the chronologically shuffled versions in both of the scenarios at accuracy above $90\%$. The results demonstrate that our proposal to use chronologically shuffled events as negative samples works favorably towards understanding texts in terms of chronology.

\textbf{Motion-to-text retrieval including shuffled texts.} We additionally conduct a more challenging task of motion-to-text retrieval using all the texts, including the chronologically incorrect versions. Given a motion, the model retrieves the most similar text from all the descriptions including those generated by shuffling the event orders. The same models from the prior experiments are used in this comparison. We conduct the motion-to-text retrieval in both of the aforementioned scenarios. Text candidates for retrieval in ``orig $\rightarrow$ event'' includes the original text and the shuffled decomposed events, whereas in ``event $\rightarrow$ event'', the candidates are the event descriptions concatenated in the correct order, as well as the wrong order. Table~\ref{tab:TMRnegret} shows the results. 
The conventional motion-language model is not capable of distinguishing texts that resemble similar events with different chronological order in either of the scenarios. On the contrary, our training scheme allows all the methods to achieve high retrieval accuracy despite having more text candidates for retrieval, even when the texts are identical except for the order, as in ``event $\rightarrow$ event'' setting. This further demonstrates that the negative samples are effective at training the models to be better aware of the chronology of events in human motion sequences.

\textbf{Fine-tuning language models.} Experiments up to this point in this paper follow the settings and protocols proposed in TMR, where language embeddings are preprocessed. As the proposed training scheme involves enhancing the descriptive ability of the motion-language models, especially in terms of language, we also test whether fine-tuning the language model used to establish the motion-language latent space in our proposed scheme has any effect on the outcome. 

Using the original model of TMR, we attach the aforementioned language models and enable fine-tuning of the language models. We then train the model by removing elements from the original model to observe the changes to the final retrieval results. We follow the same evaluation protocol in TMR to evaluate the performance of the models. Here, we rely on ``All'' criterion to assess the retrieval performance, where the entire test set is used for the retrieval task without any modification. 
We conduct the experiments in the ``orig $\rightarrow$ event'' setting. Results for other metrics, ``All with Threshold'', ``Dissimilar Subset'', ``Small batches'', and results from ``event $\rightarrow$ event'', are in the supplementary material. 

Table~\ref{tab:languagetuning} shows the results. Quite remarkably, by removing components from the original method, the proposed training scheme is able to better distinguish chronological differences in the description. Removing the VAE textual encoder especially played an important role. As chronologically inaccurate descriptions include expressions that are very similar in terms of vocabulary used, but differ significantly in terms of order, representing sentence features as a sample from a distribution likely has a detrimental effect at recognizing and comprehending chronology. We can also observe a trend that larger language models tend to perform better when performing fine-tuning. The additional textual information plays an important role for large language models in incorporating fine-grained details of chronological differences among event descriptions.
These result demonstrate that by allowing both modalities to form better alignment with each other, the proposed training scheme allows the motion-language model to establish a more robust motion-language latent space.

\begin{table}[t]
\caption{Comparison of retrieval results between the original TMR model and its variations equipped with different language encoders, which are further fine-tuned with our chronological negative samples. ``Tune'' indicates whether the language model is fine-tuned, ``VAE'' whether VAE feature is used, and ``Rec.'' whether motion decoder is used to reconstruct the original poses.}
\label{tab:languagetuning}
    \centering
    \setlength{\tabcolsep}{2pt}
    \resizebox{0.99\linewidth}{!}{
    \begin{tabular}{llccc|cccc|cccc}
    \toprule \multirow{2}{*}{ Method } &  \multirow{2}{*}{Encoder} &\multirow{2}{*}{ Tune } &\multirow{2}{*}{ VAE }& \multirow{2}{*}{ Rec. }&  \multicolumn{4}{c}{ Text-to-motion retrieval } & \multicolumn{4}{|c}{ Motion-to-text retrieval } \\
     & & & & & $\mathrm{R} @ 1 \uparrow$ & $\mathrm{R} @ 5 \uparrow$ & $\mathrm{R} @ 10 \uparrow$ & MedR $\downarrow$ & $\mathrm{R} @ 1 \uparrow$  & R@5$\uparrow$ & $\mathrm{R} @ 10 \uparrow$ & MedR $\downarrow$ \\
    \hline\hline  
   TMR \cite{petrovich2023tmr} & DistilBERT & & \cmark & \cmark & 5.82 & 21.33 & 32.76 & 25.00  & 9.76 & 24.13 & 33.23 & 23.50 \\ \cline{1-13}
    \multirow{16}{*}{ Ours }&  \multirow{4}{*}{ DistilBERT }& \cmark & \cmark & \cmark   & 5.25  & 19.96 & 30.98 & 29.00 & 8.69  & 22.45 & 31.89 & 28.00 \\
    & & \cmark & \cmark &   & 5.36  & 20.32 & 31.04 & 28.00 & 9.19  & 23.11 & 31.64 & 28.00 \\
    & & \cmark &  & \cmark  & 6.11  & 22.08 & 32.78 & 24.00 & 10.26  & 23.54 & 33.42 & 24.00 \\
    & & \cmark &  &   & 6.55 & 22.99 & 34.60 & 22.00 & 11.18  & 25.52 & 36.38 & 21.50 \\  \cline{2-13}
     &  \multirow{4}{*}{ CLIP}& \cmark & \cmark & \cmark   & 4.68 & 16.72 & 25.91 & 41.00 & 8.05  & 18.66 & 27.24 & 39.50 \\
    & & \cmark & \cmark &   & 3.99  & 16.49 & 27.44 & 32.00 & 6.57  & 19.59 & 28.44 & 31.50 \\
    & & \cmark &  & \cmark  & 4.58  & 17.91 & 27.62 & 35.00 & 8.07 & 19.80 & 28.44 & 33.50 \\
    & & \cmark &  &   & 5.57 &  20.00 & 30.06 & 28.00 & 8.69 & 22.06 & 31.14 & 27.50 \\  \cline{2-13}
     &  \multirow{4}{*}{ t5-base }& \cmark & \cmark & \cmark   & 6.39  & 22.81 & 34.63 & 23.00 & 10.56 &  26.21 & 36.09 & 22.50 \\
    & & \cmark & \cmark &   & 4.13 & 16.45 & 26.48 & 32.00 & 6.98  & 18.68 & 27.67 & 29.50 \\
    & & \cmark &  & \cmark  & 6.98  & 24.98 & 36.75 & 19.00 & 10.90  & 27.35 & 38.02 & 19.50 \\
    & & \cmark &  &   & 5.50  & 21.62 & 33.42 & 22.00 & 9.88  & 24.77 & 35.36 & 21.50 \\ \cline{2-13}
     & \multirow{4}{*}{ t5-large }& \cmark & \cmark & \cmark   & 6.57  & 22.97 & 33.92 & 24.00 & 9.74 &  25.62 & 36.59 & 22.00 \\
    & & \cmark & \cmark &   & 4.65  & 18.84 & 31.75 & 24.00 & 8.69  & 23.72 & 33.85 & 23.50 \\
    & & \cmark &  & \cmark  & \textbf{8.03} &  \textbf{26.73} & \textbf{38.98} & \textbf{17.00} & \textbf{11.72} & \textbf{28.15} & \textbf{39.23} & \textbf{17.50} \\
    & & \cmark &  &   & 6.52  & 24.16 & 36.50 & 20.00 & 10.56 &  25.91 & 36.86 & 19.50 \\

    \hline
    \end{tabular}}
    \vspace{-10pt}
\end{table}

\begin{figure}[]
\vspace{-5pt}
    \centering
    \includegraphics[width=0.95\linewidth]{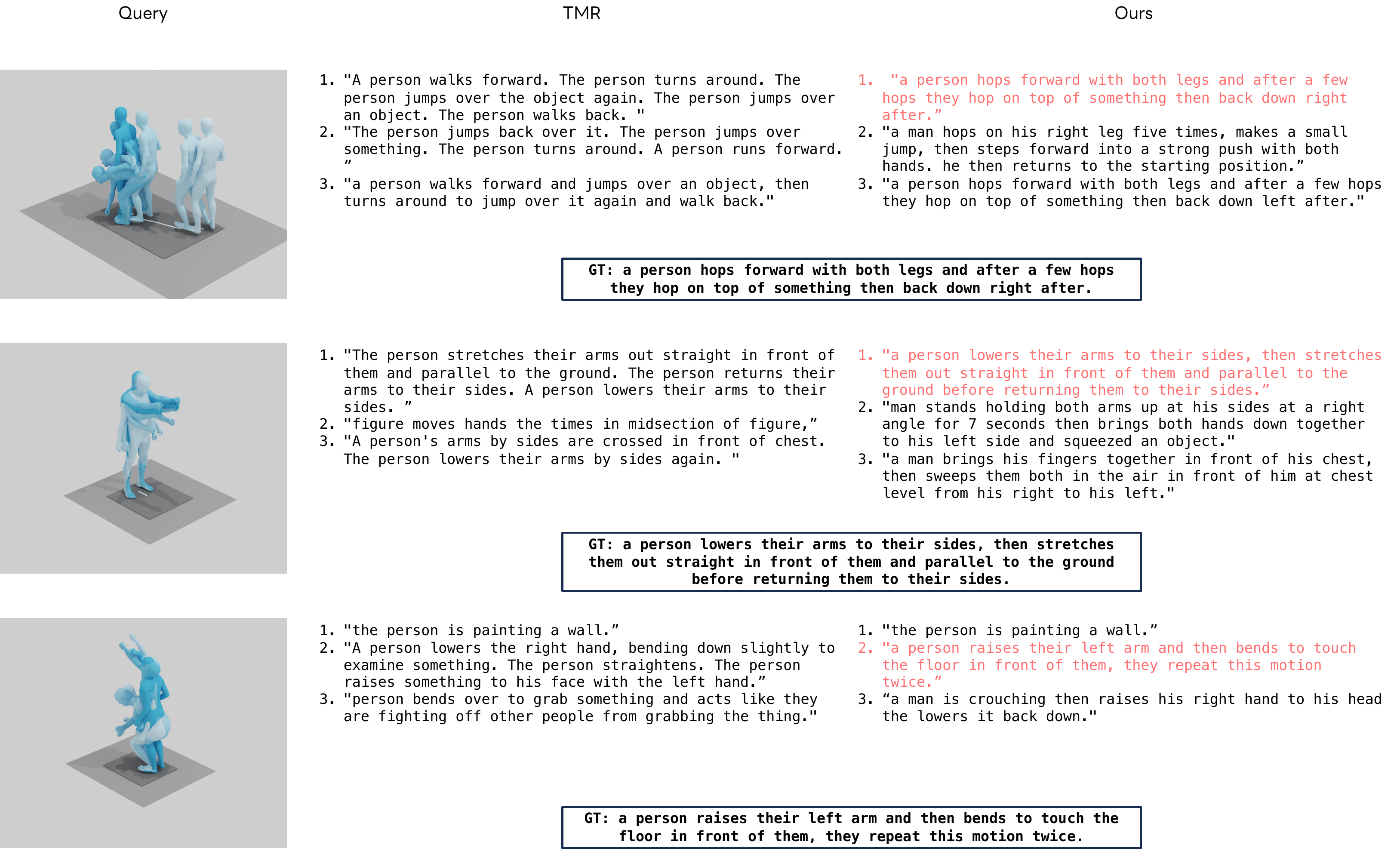}
    \caption{Comparison of retrieval results with corrupted texts using TMR and the proposed training scheme. Pink texts indicate the successfully retrieved ground truth text.}
    \label{fig:qualitative}
    \vspace{-5pt}
\end{figure}

\textbf{Qualitative Analysis.}
To further analyze the resulting motion-language model, we visualize some results from the motion-to-text retrieval including the chronologically inaccurate descriptions to demonstrate the performance of the original TMR compared to the model trained using the negative samples. 
Fig.~\ref{fig:qualitative} shows the query motions along with the retrieval results and the ground truth captions.
These concrete examples demonstrate that understanding the chronology of events can also lead to better alignment between text and motion.

\subsection{Motion Generation from Textual Descriptions}
To further evaluate the effectiveness of our proposal, we utilize the fine-tuned language model from our proposal in the task of motion generation from texts.

\textbf{Quantitative Analysis.} We first employ multiple human motion generation models to assess the descriptive capability of the language model fine-tuned through our proposal. In this analysis, we compare 3 recent models; Motiondiffuse~\cite{zhang2022motiondiffuse}, T2M-GPT~\cite{zhang2023t2m}, and ReMoDiffuse~\cite{zhang2023remodiffuse} as the base generative models. As all of these methods rely on CLIP text encoder to obtain textual features, we compare 3 variations. The first variant is the original version of each method, which uses the unaltered CLIP encoder. The second employs a CLIP encoder that is fine-tuned through TMR by enabling backpropagation to the text encoders. We note that the second variant does not use the negative textual samples to tune the text model. The last utilizes the CLIP text encoder that is fine-tuned through our proposal, in other words, the encoder is fine-tuned by also incorporating the negative chronological sample texts. Following the prior works, we evaluate the models using Frechet Inception Distance (FID), Diversity, multi-modality (MModality), Top 1-3 retrieval precision (R-Precision), and Multi-modal Distance (MM-Dist). Detailed definitions of these metrics are introduced in the supplementary material.
We conduct generation 10 times for each model, and the statistics of the results are shown in Table~\ref{tab:comptune}. As can be seen from the results, fine-tuning language models has a positive effect on the outcome of motion generation. Additionally, from the bottom rows of the table, we can observe that our proposal of using negative chronological text samples improves the metrics further in all of the generation models in this experiment. 

We further analyze the effect of different language models fine-tuned through our proposal on the generated motions. We select T2M-GPT~\cite{zhang2023t2m} as the base generation model, as the method uses a single token from the text encoder as the initial token for motion generation, which facilitates the customization. We employ the feature corresponding to the initial textual token to be used for the initial motion token of T2M-GPT. We compare the performance of different language models fine-tuned through our proposal by using negative chronological samples. As can be seen from the results in Table~\ref{tab:compt2m}, all of the models demonstrate remarkable improvement in performance over baseline methods, including one that utilizes a fine-tuned language model without the negative samples.

\begin{table}[t]
\caption{Comparison of performance of motion generation models and their variants. ``Tune'' indicates CLIP text encoders fine-tuned from text-motion retrieval tasks. ``Neg'' indicates the usage of negative chronological samples in contrastive learning.}
\label{tab:comptune}
\vspace{-5pt}    
\setlength{\tabcolsep}{2pt}
    \resizebox{0.99\linewidth}{!}{\begin{tabular}{lccccccccc}
\hline \multirow{2}{*}{ Models } & \multirow{2}{*}{ Tune } &  \multirow{2}{*}{ Neg } & \multicolumn{3}{c}{ R-Precision $\uparrow$} & \multirow{2}{*}{ FID $\downarrow$} & \multirow{2}{*}{ MM-Dist $\downarrow$} & \multirow{2}{*}{Diversity$\uparrow$}  & \multirow{2}{*}{MModality$\uparrow$}  \\
 \cline { 4 - 6 } & & & Top-1 & Top-2 & Top-3 & & & & \\
\hline Real motion & & & $0.511^{ \pm .003}$ & $0.703^{ \pm .003}$ & $0.797^{ \pm .002}$ & $0.002^{ \pm .000}$ & $2.974^{ \pm .008}$ & $9.503^{ \pm .065}$ & - \\
\hline Motiondiffuse & & & $0.486^{ \pm .005}$ & $0.681^{ \pm .003}$ & $0.783^{ \pm .003}$ & $0.651^{ \pm .025}$ & $3.079^{ \pm .011}$ & $9.466^{ \pm .110}$ & $2.204^{ \pm .086}$ \\
T2M-GPT & & & $0.489^{ \pm .004}$ & $0.677^{ \pm .003}$ & $0.774^{ \pm .003}$ & $0.116^{ \pm .005}$ & $3.287^{ \pm .013}$ & $9.771^{ \pm .104}$ & $1.710^{ \pm .061}$ \\
ReMoDiffuse & & & $0.489^{ \pm .005}$ & $0.676^{ \pm .004}$ & $0.774^{ \pm .003}$ & $0.140^{ \pm .074}$ & $3.287^{ \pm .013}$ & $9.247^{ \pm .101}$ & $2.633^{ \pm .101}$ \\
\hline
Motiondiffuse & \cmark & & $0.506^{ \pm .003}$ & $0.697^{ \pm .002}$ & $0.797^{ \pm .003}$ & $0.414^{ \pm .023}$ & $2.981^{ \pm .022}$ & $9.606^{ \pm .099}$ & $2.582^{ \pm .088}$ \\
T2M-GPT & \cmark & & $0.498^{ \pm .003}$ & $0.691^{ \pm .005}$ & $0.785^{ \pm .003}$ & $0.107^{ \pm .008}$ & $3.051^{ \pm .019}$ & $9.679^{ \pm .076}$ & $1.687^{ \pm .085}$ \\
ReMoDiffuse & \cmark & &$0.525^{ \pm .003}$ & $0.719^{ \pm .002}$ & $0.814^{ \pm .003}$ & $0.146^{ \pm .009}$ & $2.839^{ \pm .012}$ & $9.301^{ \pm .095}$ & $2.362^{ \pm .103}$ \\
\hline
Motiondiffuse & \cmark & \cmark& $0.513^{ \pm .004}$ & $0.710^{ \pm .005}$ & $0.804^{ \pm .003}$ & $0.367^{ \pm .018}$ & $3.115^{ \pm .016}$ & $9.535^{ \pm .080}$ & $2.640^{ \pm .091}$ \\
T2M-GPT & \cmark & \cmark& $0.528^{ \pm .004}$ & $0.717^{ \pm .003}$ & $0.806^{ \pm .002}$ & $0.070^{ \pm .006}$ & $2.918^{ \pm .017}$ & $9.659^{ \pm .145}$ & $1.265^{ \pm .145}$ \\
ReMoDiffuse & \cmark & \cmark& $0.525^{ \pm .005}$ & $0.719^{ \pm .006}$ & $0.813^{ \pm .004}$ & $0.116^{ \pm .010}$ & $2.881^{ \pm .011}$ & $9.248^{ \pm .121}$ & $2.449^{ \pm .125}$ \\
\hline 
\end{tabular}}
\vspace{-5pt}
\end{table}

\begin{table}[t]
\caption{Comparison of performance of motion generation with different language models using T2M-GPT as the base model. ``Tune'' indicates fine-tuning language models through backpropagating TMR, and ``Neg'' indicates using negative chronological samples. Note that the original T2M-GPT relies on CLIP encoder for the initial token.}
\label{tab:compt2m}
\setlength{\tabcolsep}{2pt}
\resizebox{0.99\linewidth}{!}{
\begin{tabular}{lccccccccc}
\hline
\multirow{2}{*}{ Encoder } &\multirow{2}{*}{ Tune }  &\multirow{2}{*}{ Neg } & \multicolumn{3}{c}{ R-Precision $\uparrow$} & \multirow{2}{*}{ FID $\downarrow$} & \multirow{2}{*}{ MM-Dist $\downarrow$} & \multirow{2}{*}{Diversity$\uparrow$}  & \multirow{2}{*}{MModality$\uparrow$}  \\
\cline { 4- 6 } & & & Top-1 & Top-2 & Top-3 & & & & \\
\hline Real motion & & & $0.511^{ \pm .003}$ & $0.703^{ \pm .003}$ & $0.797^{ \pm .002}$ & $0.002^{ \pm .000}$ & $2.974^{ \pm .008}$ & $9.503^{ \pm .065}$ & - \\
\hline 
 \multirow{2}{*}{CLIP}& & & $0.489^{ \pm .004}$ & $0.677^{ \pm .003}$ & $0.774^{ \pm .003}$ & $0.116^{ \pm .005}$ & $3.287^{ \pm .013}$ & $9.771^{ \pm .104}$ & ${1.710^{ \pm .061}}$ \\
& \cmark & & $0.498^{ \pm .003}$ & $0.691^{ \pm .005}$ & $0.785^{ \pm .003}$ & $0.107^{ \pm .008}$ & $3.051^{ \pm .019}$ & $9.679^{ \pm .076}$ & $1.687^{ \pm .085}$ \\
\cline { 1 - 10 }
DistilBERT &\cmark & \cmark & $0.528^{ \pm .007}$ & $0.717^{ \pm .008}$& $0.810^{ \pm .004}$& $0.074^{ \pm .012}$ & $2.915^{ \pm .020}$ & $9.499^{ \pm .197}$ & $1.599^{ \pm .075}$ \\
CLIP&\cmark & \cmark & $0.528^{ \pm .004}$ & $0.717^{ \pm .003}$ & $0.806^{ \pm .002}$ & $0.070^{ \pm .006}$ & $2.918^{ \pm .017}$ & $9.659^{ \pm .145}$ & $1.265^{ \pm .145}$ \\
t5base&\cmark & \cmark & ${0.530^{ \pm .005}}$ &$0.719^{ \pm .002}$& $0.810^{ \pm .003}$ & $0.087^{ \pm .009}$ & ${2.896}^{ \pm .010}$ & $9.716^{ \pm .246}$ & $1.536^{ \pm .156}$ \\
t5large&\cmark & \cmark & $0.529^{ \pm .007}$ &$0.718^{ \pm .005}$& $0.812^{ \pm .006}$ & $0.074^{ \pm .008}$ & $2.904^{ \pm .029}$ &  $9.681^{ \pm .112}$ & $1.670^{ \pm .077}$ \\
\hline 
\end{tabular}}
\vspace{-5pt}
\end{table}

To evaluate the chronological accuracy of motion generation, we take advantage of the autoregressive characteristics of T2M-GPT and compare the cumulative likelihood of returning the ground truth motion tokens when using true texts and shuffled ones as input of T2M-GPT. In this experiment, we used the CLIP model for the text encoder, with ours fine-tuned by disabling the VAE loss using the negative events. The original model returned a better likelihood for the true text inputs in only $61.9\%$ of test set motions compared to the shuffled inputs. However, the figure was $89.9\%$ for the model tuned by our method, indicating the effectiveness of considering chronology also during generation.

\begin{figure}[t]
\vspace{-5pt}
    \centering
    \includegraphics[width=1\linewidth]{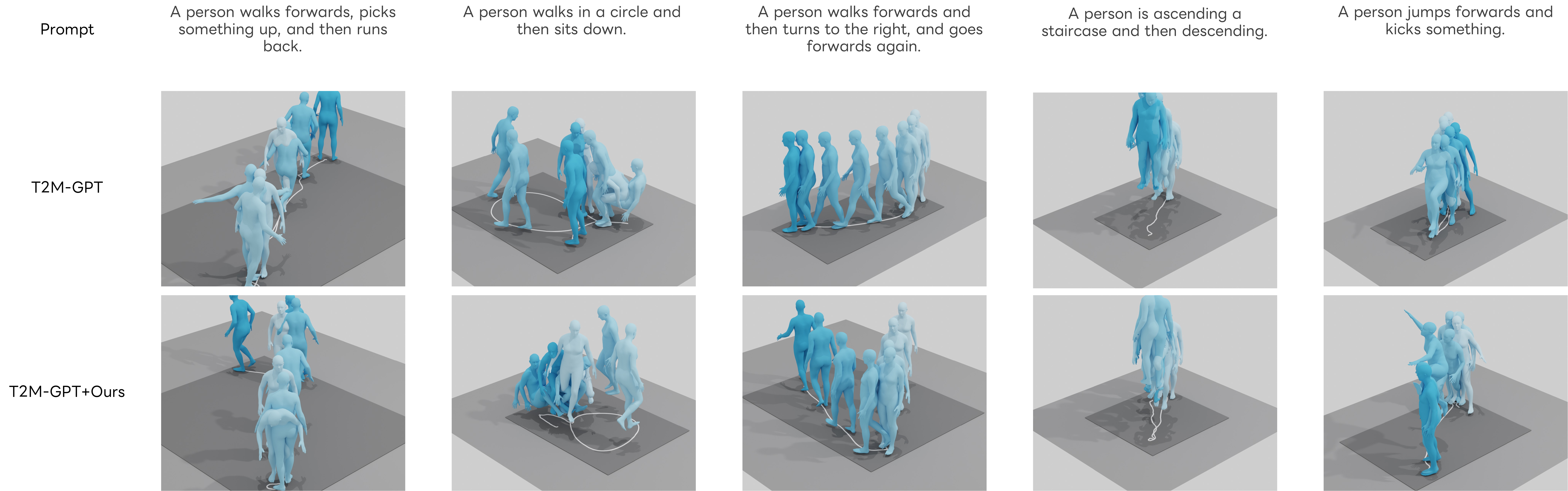}
    \caption{Comparison of generated motions. Top: T2M-GPT. Bottom: T2M-GPT with our fine-tuned t5-large encoder. Texts at the top represent the input prompts.}
    \label{fig:gent2m}
    \vspace{-10pt}
\end{figure}

\textbf{Qualitative Analysis} To qualitatively assess the generated motions, we present some of the example motions generated by the models. Fig~\ref{fig:gent2m}  shows the generated results. We base the comparison on the same model T2M-GPT~\cite{zhang2023t2m}. The top row shows the motions generated by the original T2M-GPT, and the bottom row the motions generated using T2M-GPT with the fine-tuned text-encoder trained with our proposal. 
When the prompt demands a sequence of compound actions, the model fails to fully capture all the events, for example, the person is not picking anything in the first case, and is not descending in the fourth example. There are also cases where chronology is wrong, as in the second example. On the other hand, the fine-tuned language model trained using our proposal is able to capture each event in the prompt in many of the cases.

\section{Limitations}

Although we achieved better correspondence between motion and language with our proposal in terms of chronology, we noticed that the existence of pronouns and articles in texts may provide implicit hints regarding the chronology of the events, as the original texts use definitive articles and avoid repetitive pronouns if a certain subject appears more than once in the text.

To observe the effect of these elements, we compared the CAR metric under different settings. The CAR metric for the best fine-tuned model t5-large in the original experiment was at $94.62\%$. First, we unified the articles `a, an, the' at the start of sentences and events to `The' in order to conceal the implicit information regarding when the event is taking place. This led to the CAR of $94.51\%$ for the same fine-tuned model. There are also words that points to the person in the sequence, such as `she' and `he'. When additionally replacing them with `The person' in the test set, the CAR metric for the same model fell to $81.21\%$. We discuss this issue in detail in the supplementary material.

Also, correspondence between individual words and motion is yet to be achieved, as sentences are compressed to a single feature. Exploring methods to localize the correspondence between texts and the frames of the motions is one of future works.
As the proposal mainly focused on manipulating textual descriptions, we would also like to reinforce the current proposal by augmenting motion sequences so that motion event orders can also be manipulated. 
\section{Conclusion}

We focused on the temporal element of motion and language, and uncovered that current approaches mostly overlook the factor of chronological alignment between text and motion, through a novel metric assessing event chronology.
We proposed a simple solution to achieve better correspondence between language and motion, demonstrating the importance of chronology through various experiments. We hope to inspire other research to focus more on compound actions to establish a better representation for human motion, a complex medium of data.

%
%
\bibliographystyle{splncs04}
\bibliography{main}
\clearpage
\section*{Supplementary Material}

\section{Motion Data}
\label{sec:data}
As stated in the main manuscript, we mostly follow the setting in the original paper of TMR~\cite{petrovich2023tmr}. For pose representation, we converted each frame of $J$ joints the in the motion sequence $M$ into a 263 dimensional vector consisting of ($r^{va}$, $r^{vx}$, $r^{vz}$, $r_{h}$, $\textbf{j}^p$, $\textbf{j}^v$, $\textbf{j}^r$, $\textbf{f}$), where $r^{va}$, $r^{vx}$, $r^{vz}$ and $r_{h}$ are the angular velocity around y axis, velocity in x direction, velocity in z direction and height of the root on the XZ-plane, respectively. $\textbf{j}^p \in \mathbb{R}^{3(J-1)},\textbf{j}^v \in \mathbb{R}^{3J}$ indicate the positions and velocities of the $J$ joints. $\textbf{j}^r \in \mathbb{R}^{6(J-1)}$ denotes the six-dimensional representation of rotation for each joint. $\textbf{f} \in \mathbb{R}^4$ are the contact state of the feet.

For chronological event shuffling, we select sequences that contain multiple events. In HumanML3D dataset, multiple events are detected in 13,044 out of 23,384 sequences, 811 out of 1,460 sequences, 2,677 out of 4,380 sequences in the training, validation, and test sets, respectively.


\section{Text-Motion Retrieval}
\subsection{TMR with Different Language Models}

\begin{table}[b]
\caption{Retrieval results of TMR with different language models.}
    \label{tab:TMRinit}
    \centering
    \setlength{\tabcolsep}{2pt}
    \resizebox{0.99\linewidth}{!}{
    \begin{tabular}{l|cccc|cccc}
    \toprule \multirow{2}{*}{ Models } & \multicolumn{4}{c}{ Text-to-motion retrieval } & \multicolumn{4}{|c}{ Motion-to-text retrieval } \\
     & $\mathrm{R} @ 1 \uparrow$ &  $\mathrm{R} @ 3 \uparrow$ & $\mathrm{R} @ 5 \uparrow$ & MedR $\downarrow$ & $\mathrm{R} @ 1 \uparrow$ &  $\mathrm{R} @ 3 \uparrow$ & R@5$\uparrow$ &  MedR $\downarrow$ \\
    \hline\hline DistilBERT & 5.82 & 14.85 & 21.33  & 25.00 &  9.76   & 18.13 & 24.13 & 23.50  \\
     CLIP & 4.22 & 10.58 & 15.92 &  39.00 & 6.52  & 13.37 & 17.91 &  37.50 \\
     t5-base  & 5.75 & 14.83 & 20.99 &  28.00 & 8.67  & 16.61 & 22.83 &  26.50 \\
     t5-large & 5.25  & 13.71 & 19.48 & 29.50 & 9.06  & 16.67 & 21.10 &  28.00 \\
    \hline

    \end{tabular}}
\end{table}

As the original method of TMR is equipped with DistilBERT language model to encode texts, we first observe how TMR behaves when combined with different language models. We trained each variation of TMR in a similar manner as the original model only using the original texts. Table~\ref{tab:TMRinit} shows the results. As can be seen, without training with the hard negative samples, all the variants perform similarly, showing no preference towards larger models. It is interesting to note that the CLIP encoder, which is a popular choice in motion analysis, performs the worst among all the models.

\subsection{Training with Negative Descriptions: Configuration}
\label{sec:exp}
We describe how our models are fine-tuned using the shuffled events as hard negative samples. As stated in the main manuscript, the model used in this paper follows that of TMR~\cite{petrovich2023tmr} for fairness of comparison. The model consists of two Transformer encoders, each corresponding to text and motion. Following ACTOR~\cite{petrovich2021action}, each Transformer is equipped with additional learnable parameters corresponding to $\mu$ and $\sigma$ that determine the Gaussian distribution from which the latent vectors are sampled. The dimension $d$ of the latent vectors is set at $256$. The motion decoder is the same as in ACTOR. 
\begin{table}[t]
    \caption{Comparison of retrieval results from training TMR with DistilBERT as language model using negative chronological samples under different batch size.}
    \label{tab:batch}
    \vspace{-5pt}
    \centering
    \setlength{\tabcolsep}{4pt}
    \resizebox{0.99\linewidth}{!}{
    \begin{tabular}{c|ccc|ccc|c}
    \toprule 
     & \multicolumn{7}{c}{``orig $\rightarrow$ event''} \\
    \multirow{2}{*}{Batch Size}   & \multicolumn{3}{c}{ Text-to-motion retrieval } & \multicolumn{3}{|c|}{ Motion-to-text retrieval } & \multirow{2}{*}{CAR} \\
     & $\mathrm{R} @ 1 \uparrow$ &  $\mathrm{R} @ 5\uparrow$ &  MedR $\downarrow$ & $\mathrm{R} @ 1 \uparrow$  & R@5$\uparrow$  & MedR $\downarrow$ &  \\
    \hline\hline 
    16 & 6.27   & 22.29 & 24.00 & 10.10  & 24.43  & 23.50 & 99.70  \\ 
    32  & 5.75   & 20.69 &  25.00 &  10.45 & 24.29  & 25.50 & 99.59 \\
    64  & 5.73   & 22.01 & 25.00  & 9.15  & 22.99  & 24.00 & 99.07\\
    128  & 6.04   & 21.67 &  25.00 & 9.49  & 24.68  & 23.50 & 98.73 \\
     256 & 6.07   & 22.60  & 25.00 & 9.88 & 24.91  & 24.00 & 98.36\\
     
    \hline
    \end{tabular}}
    \vspace{-5pt}
   
\end{table}

For the experiment in Fig 3 and Table 1 in the main manuscript, we followed the original method of TMR by using all the token embeddings from the textual descriptions. That is, we preprocess the text by feeding the texts into the language models, DistilBERT~\cite{sanh2019distilbert}, CLIP~\cite{radford2021learning}, t5-base, and t5-large~\cite{raffel2020exploring}. We then attach the two learnable parameters to the preprocessed token embeddings from each model and conduct the training. 

For training, the same loss from TMR~\cite{petrovich2023tmr} is used. In other words, the loss is $\mathcal{L}_{\textnormal{TMR}} = \lambda_{R}\mathcal{L}_{R}+\lambda_{KL}\mathcal{L}_{KL}+\lambda_{E}\mathcal{L}_E + \lambda_{C}\mathcal{L}_C$, where $\mathcal{L}_{R}$, $\mathcal{L}_{KL}$, $\mathcal{L}_{E}$, and $\mathcal{L}_{C}$ are reconstruction loss, KL divergence loss, embedding similarity loss, and contrastive loss, respectively. TMR employs InfoNCE~\cite{oord2018representation} in eq.(1) for the contrastive loss $\mathcal{L}_{C}$, whereas our training with chronologically negative samples uses eq.(3) for $\mathcal{L}_{C}$. The balancing weights $\lambda_{R}$, $\lambda_{KL}$, $\lambda_{E}$, $\lambda_{C}$ are set to $1$, $10^{-5}$, $10^{-5}$, and $0.1$, respectively in the experiments. 

We use the AdamW optimizer~\cite{loshchilov2018decoupled} with a learning rate of $10^{-4}$, as in TMR. We set the batch size at $32$, and are trained for 200 epochs for optimal performance. We show the effect of batch size for the ``orig $\rightarrow$ event'' scenario in Table~\ref{tab:batch}. As can be seen, despite some fluctuations, there is little difference among the results. For a fair comparison between language models, we set the maximum number of tokens to $77$, which is the limit of the CLIP language encoder. Other hyperparameters are set to be exactly the same as in TMR. The models are implemented in PyTorch, and are trained using a single Tesla A100 GPU.

\subsection{Comparison to Prior Methods}
Although TMR is the most advanced method available, we enlist the CAR metric and the corrupted motion-to-text retrieval results in ``orig $\rightarrow$ event'' setting, from methods with publicly available pretrained models, in Table~\ref{tab:prior}. For fairness, we list the results from DistilBERT for ``Ours''. We note that the language model is kept fixed in this experiment.

\begin{table}[t]
\vspace{-5pt}
    \caption{Comparison of CAR and Motion-to-text retrieval with corrupted texts with different conventional models. The language model is fixed in this setting.}
    \label{tab:prior}
    \centering
    \setlength{\tabcolsep}{4pt}
    \resizebox{0.99\linewidth}{!}{
    \begin{tabular}{lc|cccccc}
    \toprule \multirow{2}{*}{ Models } & \multirow{2}{*}{ CAR }   & \multicolumn{6}{|c}{ Motion-to-text retrieval with corrupted text} \\
      & & $\mathrm{R} @ 1 \uparrow$ & $\mathrm{R} @ 2 \uparrow$ & $\mathrm{R} @ 3 \uparrow$ & R@5$\uparrow$ & $\mathrm{R} @ 10 \uparrow$ & MedR $\downarrow$ \\
    \hline
    Guo et al. [13] & 29.32 & 0.41 & 0.62 & 0.75 & 1.25 & 2.24 & 559.25 \\
    TEMOS [30] & 62.16& 3.31 & 4.24 & 6.52 & 8.80 & 13.98 & 167.00 \\
     TMR [31] &  62.23& 7.07 & 9.65 & 12.96 & 18.09 & 26.60 & 37.50 \\\hline 
     Ours & \textbf{99.07}&  \textbf{9.03} & \textbf{11.45} & \textbf{17.11} & \textbf{22.76} & \textbf{33.14} & \textbf{24.50} \\
    \hline
    \end{tabular}}
    \vspace{-5pt}

\end{table}

\subsection{Fine-tuning Language Models}
When fine-tuning the language models to conduct experiments shown in Table 2 in the main manuscript, some modifications to TMR are required to enable the training. 

Disabling VAE functionality requires text token embeddings to be compressed into a single vector. To conduct this, we simply use the first token embedding for outputs from DistilBERT, t5-base, and t5-large. For CLIP, we extract the feature corresponding to the EoT token embedding. Instead of using these feature vectors as input to the original Transformer-based encoder of ACTOR, we instead use this as input to a simple projection head, consisting of 2 linear layers with GELU~\cite{hendrycks2016gelu} activation and layer normalization~\cite{ba2016layer}. Motion encoder is kept the same but used without the VAE functionality. The weights for losses $\lambda_{R}$, $\lambda_{E}$, $\lambda_{C}$ are kept the same. However, as the method no longer solves for the distributions, we omit the KL loss.

Disabling reconstruction loss removes the necessity of the motion decoder. We therefore remove the component. As the weight for the reconstruction loss $\lambda_{R}$ is dominant in the other settings, we instead raise the weight of the contrastive loss $\lambda_{C}$ to $1.0$. If VAE is also disabled, we also remove the KL loss, as it is unnecessary. 
We set the learning rate $10^{-5}$ for optimizing the language models to enable refinement of text embeddings, and maintain $10^{-4}$ to optimize the rest of the model. To enable conducting experiments in the same setting we use a batch size of 32, as t5 models require larger memory space. To also observe the capabilities of each language model, the maximum number of tokens is raised to $128$ to enable better expression for DistilBERT, t5-base, and t5-large. 

We note that to avoid providing additional hints regarding the chronology of events, we unified the articles ``a, the'' to ``The'' in the event descriptions, as these tend to appear at the start of a description.

\subsection{Evaluation}
We use the same evaluation metric as TMR~\cite{petrovich2023tmr}: Recall at various ranks ($R@1$, $R@2$, $\dots$) and median rank (MedR). Recall at rank $k$ indicates the percentage of cases where correct answers are returned within the top $k$ results. We select the weights from the epoch that provided the best R@1 motion-to-text retrieval accuracy on the validation set.
We follow the same evaluation protocols in TMR to evaluate the performance of the models:
\begin{itemize}
    \item \textbf{(a) All}: The entire test set is used for the retrieval task without any modification. Slight changes in the text affect this metric (e.g., person/human, walk/walking). We rely on this metric in most of the tests. 
    \item \textbf{(b) All w/ Threshold}: The entire test set is used for this retrieval task as well, except, if the text is similar to the query text above a threshold, the retrieval results are accepted as correct. The threshold is set to $0.95$ in order to remove very similar expressions.
    \item \textbf{(c) Dissimilar subset}: From the entire test set, we select 100 motion-text pairs that are far apart from each other, determined by using an approximation of the quadratic knapsack problem~\cite{aider2022branch}.
    \item \textbf{(d) Small batches}: Proposed by Guo~et al., \cite{guo2020action2motion}, this scenario selects random batches of 32 motion-text pairs, and reports the average performance of the batches.
\end{itemize}

\subsection{Full Results: Proposed Fine-tuning}

\subsubsection{Motion-to-Text Retrieval Including Corrupted Texts}

As the main paper only contained results from cases where language models were kept fixed in Table 1, we tested the fine-tuned language models trained through our proposal in the task of motion-to-text retrieval including the corrupted texts as retrieval candidates. We also conduct the CAR test using our fine-tuned models. Tables~\ref{tab:m2to2e} and~\ref{tab:m2te2e} show the results in the ``orig $\rightarrow$ event'' and ``event $\rightarrow$ event'' scenarios, respectively. For the last four rows in each table indicating models fine-tuned through our method, we select the variation that disables the VAE loss, which performs best for most models. We also enlist the numbers from TMR~\cite{petrovich2023tmr} trained with the same data (excluding the negative samples) for reference. 

As can be seen from the tables, fine-tuning the language models achieves higher performance than the non-tuned results from Table 1 in the main manuscript. The negative samples reinforce the models to differentiate accurate descriptions from the corrupted ones, even in the difficult setting of ``event $\rightarrow$ event'', where all the descriptions consist of concatenated events. A similar tendency can be seen in the CAR accuracy as well, slightly improving from the non-tuned results shown in Fig 3 in the main manuscript. As differentiating similar text with different chronological order requires richer textual representation, the larger language models perform better overall, with the best retrieval results attained with the tuned t5-large model. 

\begin{table}[t]
    \caption{Motion-to-text retrieval results with both the original and corrupted texts, and CAR accuracy in ``orig $\rightarrow$ event'' scenario. Here, the model is asked to to distinguish the original text from the corrupted text obtained by shuffling events.}
    \label{tab:m2to2e}
    \vspace{-5pt}
    \centering
    \setlength{\tabcolsep}{4pt}
    \resizebox{0.99\linewidth}{!}{
    \begin{tabular}{c|cccccc|c}
    \toprule 
    \multirow{2}{*}{Model}   &  \multicolumn{6}{|c|}{ Motion-to-text retrieval w/ corrupted events} &  \multirow{2}{*}{CAR$ \uparrow$}  \\
     & $\mathrm{R} @ 1 \uparrow$ &  $\mathrm{R} @ 2\uparrow$ &  $\mathrm{R} @ 3 \uparrow$ & $\mathrm{R} @ 5 \uparrow$  & $\mathrm{R} @ 10 \uparrow$  & MedR $\downarrow$ &  \\
    \hline\hline 
    TMR~\cite{petrovich2023tmr} & 7.76 & 10.01 & 14.51 & 20.14 & 28.54 & 33.50 & 65.45 \\ 
    \hline DistilBERT & 10.26 & 12.77 & 18.43 & 23.54 & 33.42 & 24.00 &  \textbf{99.59}\\
    CLIP  & 8.07 & 9.97 & 14.30 & 19.80 & 28.44 & 33.50 & 99.07\\
    t5-base  & 10.90 & 13.82 & 19.73 & 27.35 & 38.02 & 19.50 &  98.73 \\
     t5-large & \textbf{11.70} & \textbf{15.19} & \textbf{21.65} & \textbf{28.15} & \textbf{39.23} & \textbf{17.50} & 98.36 \\
     
    \hline
    \end{tabular}}
    \vspace{-5pt}
\end{table}

\begin{table}[t]
    \caption{Motion-to-text retrieval results with both the original and corrupted texts, and CAR accuracy in ``event $\rightarrow$ event'' scenario. Here, the model is asked to to distinguish the event descriptions concatenated in the original order from the corrupted text obtained by shuffling events.}
    \label{tab:m2te2e}
    \vspace{-5pt}
    \centering
    \setlength{\tabcolsep}{4pt}
    \resizebox{0.99\linewidth}{!}{
    \begin{tabular}{c|cccccc|c}
    \toprule 
      \multirow{2}{*}{Model}   &   \multicolumn{6}{|c|}{ Motion-to-text retrieval w/ corrupted events} & \multirow{2}{*}{CAR$ \uparrow$} \\
     & $\mathrm{R} @ 1 \uparrow$ &  $\mathrm{R} @ 2\uparrow$ &  $\mathrm{R} @ 3 \uparrow$ & $\mathrm{R} @ 5 \uparrow$  & $\mathrm{R} @ 10 \uparrow$  & MedR $\downarrow$ &  \\
    \hline\hline 
    TMR~\cite{petrovich2023tmr} & 6.61 & 8.92 & 12.73 & 17.88 & 26.53 & 35.00 & 64.29 \\ 
    \hline DistilBERT & 8.05 & 10.56 & 15.65 & 20.89 & 31.98 & 27.00&  93.69\\
    CLIP  & 6.98 & 8.58 & 12.84 & 18.04 & 26.30 & 44.25 & 92.53 \\
    t5-base  & 10.13 & 12.82 & 18.43 & 24.73 & 35.63 & \textbf{20.50} & 93.50 \\
     t5-large & \textbf{10.70} & \textbf{13.57} & \textbf{19.32} & \textbf{26.07} & \textbf{36.88} & \textbf{20.50} & \textbf{94.51}\\
     
    \hline
    \end{tabular}}
    \vspace{-5pt}
\end{table}

\subsubsection{Conventional Text-Motion Retrieval}
We then demonstrate the performance of the models fine-tuned by our proposal in the conventional retrieval task, where shuffled event descriptions do not exist in the test data. 

In the main manuscript, we only listed the results from ``(a) All'' for the ``orig $\rightarrow$ event'' scenario due to the limitation of space. 
Table~\ref{tab:fullres} shows the complete results of the fine-tuning experiment with different language models for the ``orig $\rightarrow$ event'' scenario, and Table~\ref{tab:fullres2} the results of the ``event $\rightarrow$ event'' scenario. As with the prior experiments, we unified the articles ``a, the'' to ``The'' in the event descriptions. This means that models in ``orig $\rightarrow$ event'' are tested with these rectified events, and ``event $\rightarrow$ event'' models are trained and tested with these events.  As can be seen from the results, most of the cases perform better as the components are removed when fine-tuning the language models. Similar to the previous results, larger models such as t5-base and t5-large tend to perform better overall in comparison to other models. Although there are some fluctuations, these two models perform best when the VAE loss is removed and the reconstruction loss is kept in the model. Despite performing slightly worse in most cases, as it is a more difficult scenario, the same trend can be observed from the results of ``event $\rightarrow$ event'' scenario. 

\begin{table}[t]
\caption{Retrieval results from ``event $\rightarrow$ event'' limited to sequences with multiple events. between the original TMR model and its variations equipped with different language models, which are fine-tuned with chronological negative samples. All the models are trained including single-event sequences.}
    \label{tab:multiple}
    \vspace{-5pt}
    \centering
    \setlength{\tabcolsep}{2pt}
    \resizebox{0.99\linewidth}{!}{
    \begin{tabular}{l|cccccc|cccccc}
    \toprule &  \multicolumn{6}{c}{ Text-to-motion retrieval } & \multicolumn{6}{|c}{ Motion-to-text retrieval } \\
     & $\mathrm{R} @ 1 \uparrow$ & $\mathrm{R} @ 2 \uparrow$ & $\mathrm{R} @ 3 \uparrow$ & $\mathrm{R} @ 5 \uparrow$ & $\mathrm{R} @ 10 \uparrow$ & MedR $\downarrow$ & $\mathrm{R} @ 1 \uparrow$ & $\mathrm{R} @ 2 \uparrow$ & $\mathrm{R} @ 3 \uparrow$ & R@5$\uparrow$ & $\mathrm{R} @ 10 \uparrow$ & MedR $\downarrow$ \\
    \hline\hline 
    TMR \cite{petrovich2023tmr}   & 8.33 & 16.77 & 22.86 & 30.74 & 42.32 & 15.00 & 13.90 & 17.93 & 24.54 & 31.57 & 43.29 & 14.50 \\ \hline
      DistilBERT& 9.15 & 16.40 & 21.63 & 29.29 & 42.73 & 15.00 & 12.10 & 15.88 & 22.34 & 29.25 & 41.05 & 15.50  \\
     CLIP   &  6.43 & 11.92 & 15.50 & 21.59 & 32.01 & 27.00 & 9.64 & 11.88 & 16.88 & 22.08 & 30.59 & 38.50  \\
    t5-base  & \textbf{10.16} & 18.98 & 24.62 & 32.98 & 47.14 & 12.00 & 14.23 & 18.27 & 26.04 & 33.10 & 45.05 & 13.50 \\
    t5-large & 9.94 & \textbf{19.20} & \textbf{25.14} & \textbf{34.14} & \textbf{48.60} & \textbf{11.00} & \textbf{15.69} & \textbf{19.50} & \textbf{27.12} & \textbf{35.34} & \textbf{47.67} & \textbf{12.00}  \\
    \hline
\end{tabular}}
    \vspace{-5pt}
\end{table}

\begin{table}[t]
    \caption{Motion-to-text retrieval results with both the original and corrupted texts in ``event $\rightarrow$ event'' scenario limited to sequences with multiple events. Here, the model is asked to to distinguish the event descriptions concatenated in the original order from the corrupted text obtained by shuffling events.}
    \label{tab:limitm2t}
    \vspace{-5pt}
    \centering
    \setlength{\tabcolsep}{4pt}
    \resizebox{0.9\linewidth}{!}{
    \begin{tabular}{l|cccccc}
    \toprule 
      \multirow{2}{*}{Model}   &   \multicolumn{6}{|c}{ Motion-to-text retrieval w/ corrupted events}  \\
     & $\mathrm{R} @ 1 \uparrow$ &  $\mathrm{R} @ 2\uparrow$ &  $\mathrm{R} @ 3 \uparrow$ & $\mathrm{R} @ 5 \uparrow$  & $\mathrm{R} @ 10 \uparrow$  & MedR $\downarrow$  \\
    \hline\hline 
    TMR~\cite{petrovich2023tmr} & 8.44 & 11.62 & 16.59 & 22.23 & 32.95 & 26.00  \\ 
    \hline DistilBERT & 10.65 & 14.42 & 20.32 & 27.19 & 38.40 & 20.50\\
    CLIP  & 8.93 & 10.98 & 15.69 & 20.84 & 28.88 & 42.50 \\
    t5-base  & 13.11 & 16.77 & 23.98 & 31.12 & 43.00 & 15.00  \\
     t5-large & \textbf{14.49} & \textbf{18.38} & \textbf{25.18} & \textbf{33.10} & \textbf{44.94} & \textbf{14.00} \\
    \hline
    \end{tabular}}
    \vspace{-5pt}
\end{table}
\subsubsection{Test Results Limited to Multiple Events}
We further investigate the effectiveness of our proposal by focusing on the retrieval results of the 2677 sequences with multiple events in the test set. Table~\ref{tab:multiple} shows the retrieval results from TMR and the models fine-tuned with our strategy for ``(a) All'' evaluation protocol, and Table~\ref{tab:limitm2t} demonstrates the results from the task of motion-to-text retrieval including the corrupted texts as retrieval candidates. The bottom 4 rows from our strategy are the variant that does not use the VAE loss for fine-tuning the language model. Similar to other experiments, the proposed method improves the larger language models in terms of retrieval accuracy. The margin of improvement of the proposal becomes larger when the task is more demanding in terms of chronological understanding, as can be seen in the results from Table~\ref{tab:limitm2t}, where chronological information is required to distinguish correct texts from corrupted ones.

\subsection{Effect of Articles and Pronouns}

In the main manuscript, we noted that the pronouns and articles may provide additional hints with regards to the chronology of events. To observe whether these elements are sufficient to determine if the order of events is correct or not, we design a simple test by training a binary classifier that learns to separate event descriptions concatenated in the original order and the shuffled descriptions. In other words, we examine whether the chronology can be determined solely from texts. We design the classifier to first encode the features from the language models in the same manner as TMR~\cite{petrovich2023tmr} without the VAE loss to achieve maximal descriptiveness. We then place a layer that receives the embedded text features and outputs the likelihood of two classes, whether they are accurate or corrupt. We train this baseline classifier under the same aforementioned protocol in Section~\ref{sec:exp} using the events concatenated in the correct order, and the shuffled ones as corrupt versions of the training set. 

We prepare 3 cases to determine the effect of such articles and pronouns. ``No Edit'': This is the case where the decomposed events are used as they are. In other words, events concatenated in the correct order are used as the positive samples, and those shuffled as negatives. No other editing is applied to the texts.  ``Article'': This indicates cases where indefinite articles are converted to definite one, as indefinite articles tend to appear at the initial event. After replacing ``A'' with ``The'', the events are treated the same as in the previous case.  ``Pronoun'': This is the case where variations of pronouns addressing a person, such as ``a figure, a person'' is unified to ``The person''. The rectified events are then processed in the same manner as the prior cases. All models are trained with data processed under the same protocol, and tested with the similarly processed data. During testing, baseline models are randomly provided with texts of either events in the original order or shuffled events, and are asked to classify whether it is shuffled.

\begin{table}[t]
    \vspace{-5pt}
    \caption{Comparison of baseline classifier accuracy and CAR of fine-tuned language models under different variations in ``event $\rightarrow$ event'' scenario. Baseline is trained to distinguish event descriptions concatenated in the original order from the shuffled ones, only from features of the corresponding encoded event descriptions.}
    \label{tab:finecar}
    \centering
    \setlength{\tabcolsep}{4pt}
    \begin{tabular}{l|c|ccc}
    \toprule {Encoder} & Method& No Edit & Article & Pronoun \\
    \hline \multirow{2}{*}{ DistilBERT} & Baseline  & 87.22 & 76.91 & 72.77 \\
     & Ours & \textbf{93.54} & \textbf{93.69} & \textbf{85.51}\\
     \hline \multirow{2}{*}{ CLIP} & Baseline &83.08 & 72.10& 70.94\\
     & Ours & \textbf{92.15}  & \textbf{92.53} & \textbf{82.11}\\

     \hline \multirow{2}{*}{ t5-base} & Baseline & 85.84 & 73.70  & 67.46\\
     & Ours & \textbf{93.50} & \textbf{93.50} & \textbf{84.58}\\

     \hline \multirow{2}{*}{ t5-large} & Baseline & 85.58 &  72.10  & 69.51\\
     & Ours &  \textbf{94.62} & \textbf{94.51} & \textbf{81.21}\\
     \hline
    \end{tabular}
    \vspace{-5pt}
\end{table}

Table~\ref{tab:finecar} shows the results from the analysis. The results from ``Baseline-No Edit'' demonstrate that there do definitely exist some leakage of chronological information solely from textual descriptions, as the baseline classifier is able to distinguish correct descriptions in approximately $85\%$ of the cases, regardless of the language model. However, our method is able to outperform the baseline in all the cases, demonstrating the ability of the proposal to extract chronological information from not only textual information, but from motion as well. 

Our strategy to stop the leakage by replacing indefinite articles with definite ones has a remarkable impact on the baseline classifier, as the accuracy drops by more than $10\%$, despite being trained with the same rectification. However, models trained through our strategy tend to perform similarly to the unedited version, displaying the power of training with negative samples. 

The most difficult case was observed when replacing pronouns indicating the actor in the sequence as ``The person''. The baseline model deteriorated further, while the model refined by our method also regressed. However, the proposed method still outperforms the baseline with all the language models.

We note that all these cases, including the baseline, still outperform the CAR results from TMR~\cite{petrovich2023tmr} in Fig. 3 in the main manuscript. This demonstrates that the prior methods are not able to take full advantage of the hints existing in the texts, indicating the necessity to delve further into the language-based aspects of motion-language models.

 There are also possibilities of other leakages of chronological information. For example, objects that a person in the sequence is interacting with, can be mentioned with a pronoun such as ``it'' as well. These could be functioning as additional information. However, as covering all such criteria requires additional analysis from the linguistics perspective, as a recent research~\cite{hsieh2024sugarcrepe} has done in the image domain,  we leave this as future work.
 
\begin{figure}[t]
    \centering
    \includegraphics[width=1\linewidth]{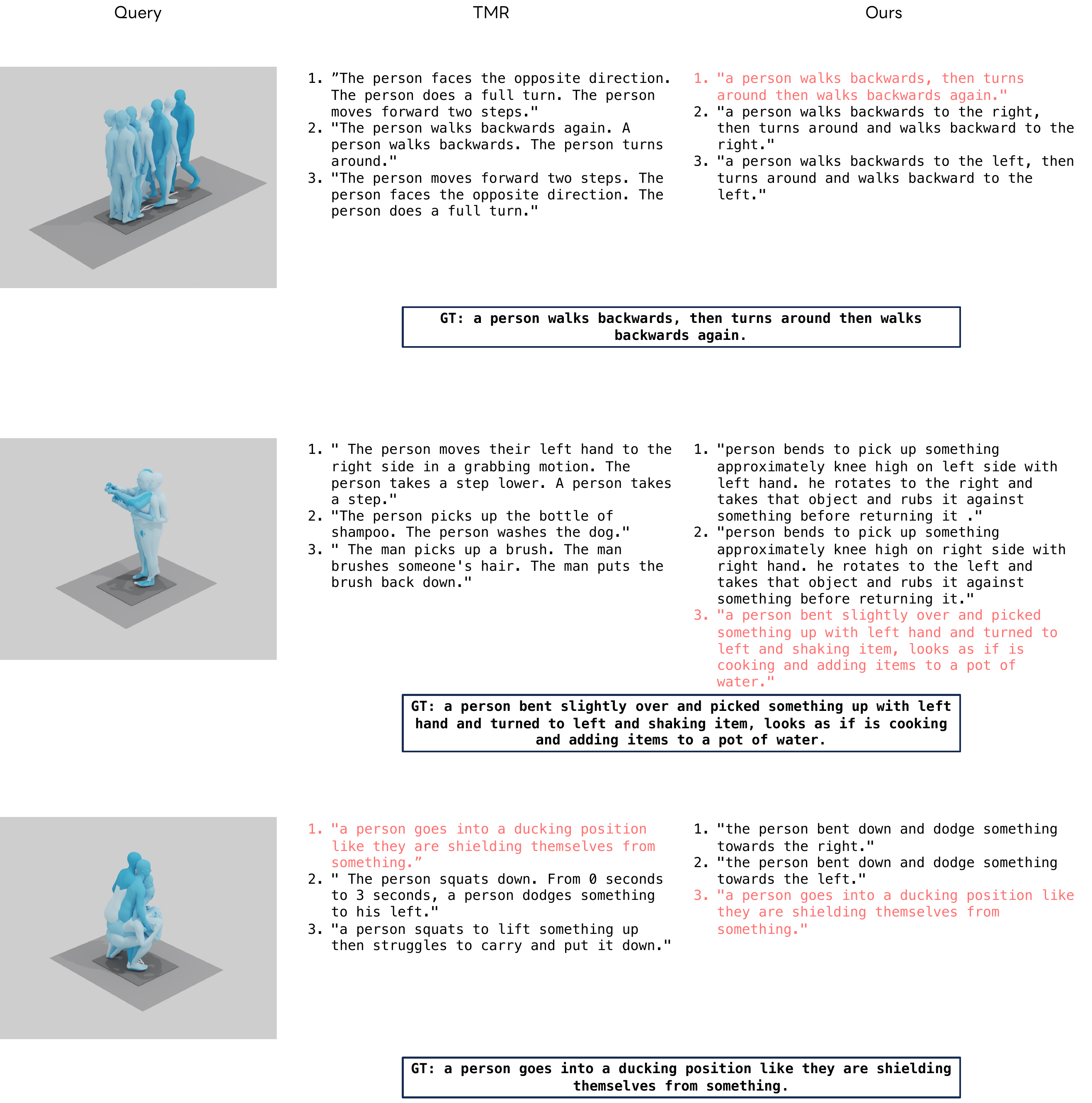}
    \caption{Additional retrieval results with corrupted texts using TMR and the proposed training scheme. Pink texts indicate the successfully retrieved ground truth text.}
    \label{fig:supqual}
\end{figure}

\subsection{Additional Qualitative Results}
We further introduce visualizations of examples from the experimental results. Fig.~\ref{fig:supqual} presents examples from TMR and results from our hard negative training. As in the main manuscript, we select these examples from the motion-to-text retrieval experiment including the shuffled texts. The setting is therefore identical to Table 3 and Fig. 6 in the main manuscript.

In the first example, all three texts retrieved with our method describe a similar event of the person going backward and then turning to go backward again. However, TMR fails to capture the chronology as well as the direction of the movement, indicating forward movement in some captions.  The second example contains an ambiguous motion of picking objects as if the person is cooking something. The proposed training scheme is able to capture the characteristics of the motion and retrieve the correct answer at the third rank. On the contrary, TMR is only capable of capturing the feature for the movement of picking something up, therefore retrieves various activities that include the motion of picking. The third case is where TMR has a better result than the proposed scheme. TMR is able to retrieve the correct text at rank 1, while the proposal is only able to pick the correct text as the third most likely text. However, all the texts in the results from the proposal resemble similar activity, whereas TMR returns a different activity that involves lifting something up as the third most similar text.

\section{Generation}

\subsection{Evaluation}
Following the prior works on human motion generation, we evaluated the generation models using the following criteria:
\begin{itemize}
    \item \textbf{(1) Top 1-3 retrieval precision (R-Precision)}: For each generated motion, its ground truth and 31 random descriptions are selected to form a pool These are ranked by the distance between the generated motion feature and the text features. The average accuracy at rank 1, 2, and 3 are measured.
    \item \textbf{(2) Frechet Inception Distance (FID)}: The distance between the feature distribution of the generated motions and that of the actual motions. The identical pretrained model as prior methods is used to extract the features.
    \item \textbf{(3) Multi-modal Distance (MM-Dist)}: Multi-modal distance measures the average distance between the generated motion features and their corresponding text features.
    \item \textbf{(4) Diversity}: Diversity is the variance of generated motion features across all the samples.
    \item \textbf{(5) Multi-modality (MModality)}: For each text, 10 pairs of motions are generated, and the average value of the mean distance of the pairs is recorded as Multi-modality.

\end{itemize}

\section{Motion Visualization}
 As motion can be difficult to observe in still images, we attach a demonstration video including some of the results from the motion-to-text retrieval including the corrupted texts, and the results from motion generation on HumanML3D dataset. The examples show the event chronology is accurately reflected in the results obtained through models tuned by the proposed strategy.

\section{Sample Code}
The code will be released at \href{https://github.com/kentfuji/ChronAccRet}{https://github.com/kentfuji/ChronAccRet}. We provide the training codes for building the motion-language model, and test codes to measure CAR, as well as the conventional evaluation metrics. We refer the readers to the README file in the code to run the experiments. 

\begin{table}[t]
\caption{Comparison of ``orig $\rightarrow$ event'' retrieval results between the original TMR model and its variations with different language models, which are fine-tuned with chronological negative samples. ``Tune'' indicates whether the language model is fine-tuned, ``VAE'' whether VAE feature is used, and ``Recon.'' whether a decoder reconstructs the poses.}
    \label{tab:fullres}
    \vspace{-5pt}
    \centering
    \setlength{\tabcolsep}{4pt}
    \resizebox{0.99\linewidth}{!}{
    \begin{tabular}{llccc|cccccc|cccccc}
    \toprule \multirow{2}{*}{ Protocol } & \multirow{2}{*}{ Method }&\multirow{2}{*}{ Tune } &\multirow{2}{*}{ VAE }& \multirow{2}{*}{ Recon. }&  \multicolumn{6}{c}{ Text-to-motion retrieval } & \multicolumn{6}{|c}{ Motion-to-text retrieval } \\
     & & & & & $\mathrm{R} @ 1 \uparrow$ & $\mathrm{R} @ 2 \uparrow$ & $\mathrm{R} @ 3 \uparrow$ & $\mathrm{R} @ 5 \uparrow$ & $\mathrm{R} @ 10 \uparrow$ & MedR $\downarrow$ & $\mathrm{R} @ 1 \uparrow$ & $\mathrm{R} @ 2 \uparrow$ & $\mathrm{R} @ 3 \uparrow$ & R@5$\uparrow$ & $\mathrm{R} @ 10 \uparrow$ & MedR $\downarrow$ \\
    \hline\hline \multirow{17}{*}{ (a) All } 
    & TMR \cite{petrovich2023tmr} &  & \cmark & \cmark & 5.82 & 11.02 & 14.85 & 21.33 & 32.76 & 25.00 & 9.76 & 12.34 & 18.13 & 24.13 & 33.23 & 23.50 \\ \cline{2-17}
    &  \multirow{4}{*}{ DistilBERT~\cite{sanh2019distilbert} }& \cmark & \cmark & \cmark   & 5.25 & 9.92 & 13.46 & 19.96 & 30.98 & 29.00 & 8.69 & 11.09 & 16.51 & 22.45 & 31.89 & 28.00 \\
    & & \cmark & \cmark &   & 5.36 & 10.83 & 14.37 & 20.32 & 31.04 & 28.00 & 9.19 & 11.27 & 17.15 & 23.11 & 31.64 & 28.00 \\
    & & \cmark &  & \cmark  & 6.11 & 11.25 & 15.74 & 22.08 & 32.78 & 24.00 & 10.26 & 12.77 & 18.43 & 23.54 & 33.42 & 24.00 \\
    & & \cmark &  &   & 6.55 & 12.25 & 16.04 & 22.99 & 34.60 & 22.00 & 11.18 & 13.57 & 19.37 & 25.52 & 36.38 & 21.50 \\  \cline{2-17}
     &  \multirow{4}{*}{ CLIP \cite{radford2021learning} }& \cmark & \cmark & \cmark   & 4.68 & 8.78 & 11.79 & 16.72 & 25.91 & 41.00 & 8.05 & 9.97 & 14.46 & 18.66 & 27.24 & 39.50 \\
    & & \cmark & \cmark &   & 3.99 & 8.14 & 11.04 & 16.49 & 27.44 & 32.00 & 6.57 & 8.55 & 14.67 & 19.59 & 28.44 & 31.50 \\
    & & \cmark &  & \cmark  & 4.58 & 8.67 & 12.00 & 17.91 & 27.62 & 35.00 & 8.07 & 9.97 & 14.30 & 19.80 & 28.44 & 33.50 \\
    & & \cmark &  &   & 5.57 & 10.61 & 13.87 & 20.00 & 30.06 & 28.00 & 8.69 & 10.77 & 16.42 & 22.06 & 31.14 & 27.50 \\  \cline{2-17}
     &  \multirow{4}{*}{ t5-base \cite{raffel2020exploring} }& \cmark & \cmark & \cmark   & 6.39 & 12.73 & 16.56 & 22.81 & 34.63 & 23.00 & 10.56 & 13.41 & 19.64 & 26.21 & 36.09 & 22.50 \\
    & & \cmark & \cmark &   & 4.13 & 8.23 & 10.65 & 16.45 & 26.48 & 32.00 & 6.98 & 8.46 & 13.55 & 18.68 & 27.67 & 29.50 \\
    & & \cmark &  & \cmark  & 6.98 & 12.82 & 18.02 & 24.98 & 36.75 & 19.00 & 10.90 & 13.82 & 19.73 & 27.35 & 38.02 & 19.50 \\
    & & \cmark &  &   & 5.50 & 10.52 & 14.60 & 21.62 & 33.42 & 22.00 & 9.88 & 12.18 & 17.86 & 24.77 & 35.36 & 21.50 \\ \cline{2-17}
     & \multirow{4}{*}{ t5-large \cite{raffel2020exploring} }& \cmark & \cmark & \cmark   & 6.57 & 12.20 & 16.15 & 22.97 & 33.92 & 24.00 & 9.74 & 12.57 & 19.34 & 25.62 & 36.59 & 22.00 \\
    & & \cmark & \cmark &   & 4.65 & 9.58 & 13.12 & 18.84 & 31.75 & 24.00 & 8.69 & 11.02 & 17.22 & 23.72 & 33.85 & 23.50 \\
    & & \cmark &  & \cmark  & \textbf{8.03} & \textbf{14.51} & \textbf{18.84} & \textbf{26.73} & \textbf{38.98} & \textbf{17.00} & \textbf{11.72} & \textbf{15.19} & \textbf{21.65} & \textbf{28.15} & \textbf{39.23} & \textbf{17.50} \\
    & & \cmark &  &   & 6.52 & 11.98 & 16.56 & 24.16 & 36.50 & 20.00 & 10.56 & 13.25 & 19.73 & 25.91 & 36.86 & 19.50 \\
    
    \hline \multirow{17}{*}{ (b) All w/ threshold } 
    
    & TMR \cite{petrovich2023tmr} & & \cmark &  \cmark& 12.32 & 16.79 & 22.99 & 30.18 & 42.34 & 15.00 & 13.64 & 15.99 & 22.47 & 28.95 & 38.44 & 18.00  \\ \cline{2-17}
    &  \multirow{4}{*}{ DistilBERT~\cite{sanh2019distilbert} }& \cmark & \cmark & \cmark   & 10.99 & 14.32 & 20.07 & 27.01 & 39.07 & 19.00 & 12.07 & 14.12 & 20.96 & 27.60 & 37.34 & 21.50 \\
    & & \cmark & \cmark &   & 13.09 & 15.67 & 21.49 & 27.97 & 38.48 & 19.50 & 12.89 & 13.48 & 21.76 & 27.21 & 35.47 & 23.00 \\
    & & \cmark &  & \cmark  & 13.00 & 16.83 & 23.79 & 30.00 & 41.42 & 16.00 & 14.14 & 16.20 & 23.08 & 28.42 & 38.57 & 19.50 \\
    & & \cmark &  &   & 13.21 & 16.77 & 23.22 & 31.68 & 42.95 & 15.00 & 15.01 & 17.22 & 24.06 & 30.34 & 41.29 & 17.50 \\  \cline{2-17}
     &  \multirow{4}{*}{ CLIP \cite{radford2021learning} }& \cmark & \cmark & \cmark   & 10.29 & 13.48 & 18.77 & 23.95 & 33.42 & 28.00 & 11.18 & 13.18 & 18.82 & 23.29 & 32.53 & 31.50 \\
    & & \cmark & \cmark &   & 10.33 & 13.46 & 18.41 & 24.84 & 36.04 & 23.00 & 10.33 & 10.72 & 18.86 & 24.20 & 32.96 & 25.50 \\
    & & \cmark &  & \cmark  & 9.79 & 13.39 & 19.30 & 25.41 & 35.17 & 22.00 & 11.52 & 13.34 & 19.07 & 25.36 & 34.24 & 25.50 \\
    & & \cmark &  &   & 12.23 & 16.20 & 21.15 & 28.22 & 39.23 & 18.50 & 12.45 & 14.03 & 20.80 & 26.53 & 35.79 & 23.00 \\  \cline{2-17}
     &  \multirow{4}{*}{ t5-base \cite{raffel2020exploring} }& \cmark & \cmark & \cmark   & 13.89 & 17.72 & 23.59 & 30.13 & 43.16 & 15.00 & 14.30 & 16.29 & 24.02 & 30.70 & 40.69 & 17.50 \\
    & & \cmark & \cmark &   & 10.54 & 15.90 & 20.10 & 27.49 & 38.07 & 20.00 & 10.08 & 10.70 & 17.31 & 22.31 & 31.09 & 25.00 \\
    & & \cmark &  & \cmark  & \textbf{15.44} & \textbf{20.87} & 27.40 & 34.99 & 47.92 & \textbf{11.00} & 14.99 & 17.93 & 24.95 & 32.62 & 42.84 & 15.00 \\
    & & \cmark &  &   & 11.86 & 15.17 & 22.51 & 30.43 & 43.07 & 14.00 & 13.48 & 14.92 & 21.88 & 28.97 & 39.28 & 17.50 \\ \cline{2-17}
     & \multirow{4}{*}{ t5-large \cite{raffel2020exploring} }& \cmark & \cmark & \cmark   & 13.21 & 17.68 & 24.16 & 31.23 & 42.52 & 15.00 & 13.32 & 15.81 & 23.56 & 30.29 & 41.65 & 17.00 \\
    & & \cmark & \cmark &   & 11.06 & 15.31 & 20.85 & 27.58 & 41.33 & 15.00 & 12.43 & 13.05 & 21.26 & 27.85 & 37.75 & 19.00  \\
    & & \cmark &  & \cmark  & 15.19 & 20.55 & \textbf{27.44} & \textbf{35.95} & \textbf{48.59} & \textbf{11.00} & \textbf{15.53} & \textbf{18.34} & \textbf{26.07} & \textbf{33.30} & \textbf{44.18} & \textbf{14.00} \\
    & & \cmark &  &   & 13.53 & 18.91 & 27.05 & 35.01 & 47.58 & 12.00 & 14.94 & 16.74 & 24.13 & 30.45 & 41.20 & 15.50 \\

    \hline \multirow{17}{*}{ (c) Dissimilar subset } 
    & TMR \cite{petrovich2023tmr} & & \cmark & \cmark & 48.00 & 71.00 & 79.00 & 86.00 & 93.00 & 2.00 & 52.00 & 73.00 & 83.00 & 86.00 & 91.00 & 1.00 \\ \cline{2-17}
    &  \multirow{4}{*}{ DistilBERT~\cite{sanh2019distilbert} }& \cmark & \cmark & \cmark   & 45.00 & 66.00 & 73.00 & 80.00 & 90.00 & 2.00 & 50.00 & 70.00 & 76.00 & 81.00 & 88.00 & 1.50 \\
    & & \cmark & \cmark &   & 52.00 & 63.00 & 69.00 & 76.00 & 82.00 & \textbf{1.00} & 50.00 & 64.00 & 74.00 & 80.00 & 83.00 & 1.50 \\
    & & \cmark &  & \cmark  & 52.00 & 65.00 & 76.00 & 84.00 & 89.00 & \textbf{1.00} & 49.00 & 73.00 & 79.00 & 85.00 & 90.00 & 2.00 \\
    & & \cmark &  &   & 53.00 & 70.00 & 77.00 & 86.00 & 88.00 & \textbf{1.00} & 53.00 & 67.00 & 77.00 & 87.00 & 88.00 & \textbf{1.00} \\  \cline{2-17}
     &  \multirow{4}{*}{ CLIP \cite{radford2021learning} }& \cmark & \cmark & \cmark  & 48.00 & 67.00 & 72.00 & 79.00 & 89.00 & 2.00 & 45.00 & 61.00 & 73.00 & 80.00 & 87.00 & 2.00 \\
    & & \cmark & \cmark &   & 50.00 & 68.00 & 70.00 & 81.00 & 87.00 & 1.50 & 52.00 & 66.00 & 72.00 & 80.00 & 89.00 & \textbf{1.00} \\
    & & \cmark &  & \cmark  & 46.00 & 66.00 & 71.00 & 79.00 & 87.00 & 2.00 & 40.00 & 69.00 & 73.00 & 79.00 & 89.00 & 2.00 \\
    & & \cmark &  &   & 54.00 & 70.00 & 79.00 & 86.00 & 91.00 & \textbf{1.00} & 57.00 & 70.00 & 78.00 & 85.00 & 92.00 & \textbf{1.00} \\  \cline{2-17}
     &  \multirow{4}{*}{ t5-base \cite{raffel2020exploring} }& \cmark & \cmark & \cmark   & \textbf{56.00} & \textbf{76.00} & 81.00 & \textbf{87.00} & 93.00 & \textbf{1.00} & 55.00 & 76.00 & 83.00 & 88.00 & 90.00 & \textbf{1.00} \\
    & & \cmark & \cmark &   & 46.00 & 66.00 & 76.00 & 85.00 & \textbf{95.00} & 2.00 & 54.00 & 69.00 & 76.00 & 88.00 & \textbf{95.00} & \textbf{1.00} \\
    & & \cmark &  & \cmark  & 55.00 & 72.00 & 76.00 & 85.00 & 94.00 & \textbf{1.00} & 55.00 & 72.00 & 79.00 & 87.00 & \textbf{95.00} & \textbf{1.00} \\
    & & \cmark &  &   & 48.00 & 71.00 & 78.00 & 86.00 & 94.00 & 2.00 & 51.00 & 70.00 & 81.00 & \textbf{89.00} & 94.00 & \textbf{1.00} \\ \cline{2-17}
     & \multirow{4}{*}{ t5-large \cite{raffel2020exploring} }& \cmark & \cmark & \cmark   & 53.00 & 68.00 & 75.00 & 86.00 & 92.00 & \textbf{1.00} & 54.00 & 71.00 & 78.00 & 87.00 & 92.00 & \textbf{1.00} \\
    & & \cmark & \cmark &   & 54.00 & 72.00 & 75.00 & 82.00 & 88.00 & \textbf{1.00} & 55.00 & 70.00 & 75.00 & 82.00 & 89.00 & \textbf{1.00} \\
    & & \cmark &  & \cmark  & 53.00 & 75.00 & \textbf{82.00} & \textbf{87.00} & 93.00 & \textbf{1.00} & \textbf{59.00} & \textbf{79.00} & \textbf{87.00} & \textbf{89.00} & 93.00 & \textbf{1.00} \\
    & & \cmark &  &   & 55.00 & 75.00 & 80.00 & 85.00 & 93.00 & \textbf{1.00} & 55.00 & 75.00 & 80.00 & 87.00 & 93.00 & \textbf{1.00} \\

    \hline \multirow{17}{*}{ (d) Small batches } 
    & TMR \cite{petrovich2023tmr} & &\cmark &\cmark& 70.19 & 83.74 & 88.82 & 93.18 & 96.83 & 1.00 & 70.64 & 84.65 & 89.46 & 93.32 & 96.67 & 1.00 \\ \cline{2-17}
    &  \multirow{4}{*}{ DistilBERT~\cite{sanh2019distilbert} }& \cmark & \cmark & \cmark   & 66.77 & 80.52 & 85.95 & 90.49 & 94.64 & 1.02 & 67.93 & 81.55 & 86.47 & 90.90 & 94.57 & 1.03 \\
    & & \cmark & \cmark &   & 65.74 & 77.99 & 82.85 & 87.02 & 91.01 & 1.03 & 66.26 & 79.15 & 83.07 & 87.14 & 90.99 & 1.05 \\
    & & \cmark &  & \cmark  & 68.20 & 82.34 & 87.29 & 91.10 & 94.73 & 1.02 & 69.69 & 82.94 & 87.39 & 90.92 & 94.39 & 1.03 \\
    & & \cmark &  &   & 70.10 & 82.69 & 87.07 & 91.13 & 94.32 & 1.01 & 70.19 & 81.98 & 86.66 & 90.69 & 94.23 & 1.01 \\  \cline{2-17}
     &  \multirow{4}{*}{ CLIP \cite{radford2021learning} }& \cmark & \cmark & \cmark   & 60.36 & 75.36 & 81.57 & 87.52 & 93.50 & 1.12 & 60.83 & 75.41 & 81.73 & 87.64 & 93.57 & 1.07 \\
    & & \cmark & \cmark &   & 65.42 & 79.43 & 84.88 & 89.01 & 93.41 & 1.01 & 65.53 & 79.77 & 85.17 & 89.14 & 92.91 & 1.04 \\
    & & \cmark &  & \cmark  & 63.16 & 77.83 & 83.90 & 89.01 & 94.64 & 1.06 & 64.12 & 78.97 & 84.60 & 89.83 & 95.23 & 1.07 \\
    & & \cmark &  &   & 67.97 & 81.34 & 86.61 & 91.15 & 95.53 & 1.02 & 67.66 & 81.93 & 87.00 & 91.10 & 95.44 & 1.01 \\  \cline{2-17}
     &  \multirow{4}{*}{ t5-base \cite{raffel2020exploring} }& \cmark & \cmark & \cmark   & 69.84 & 82.96 & 88.57 & 92.86 & 96.51 & 1.01 & 71.08 & 84.26 & 89.21 & 92.88 & 96.49 & \textbf{1.00} \\
    & & \cmark & \cmark &   & 67.84 & 84.24 & 89.80 & 94.71 & 97.97 & 1.01 & 69.53 & 84.65 & 90.65 & 95.00 & 97.99 & 1.02 \\
    & & \cmark &  & \cmark  & \textbf{75.14} & 87.86 & \textbf{92.40} & \textbf{95.92} & \textbf{98.36} & \textbf{1.00} & 74.11 & 87.96 & \textbf{92.81} & \textbf{96.10} & \textbf{98.31} & \textbf{1.00} \\
    & & \cmark &  &   & 72.19 & 86.22 & 90.94 & 94.50 & 97.45 & \textbf{1.00} & 72.54 & 86.86 & 91.20 & 94.59 & 97.58 & \textbf{1.00} \\ \cline{2-17}
     & \multirow{4}{*}{ t5-large \cite{raffel2020exploring} }& \cmark & \cmark & \cmark   & 70.69 & 83.90 & 88.69 & 93.07 & 96.08 & 1.02 & 71.56 & 84.79 & 89.21 & 93.04 & 96.08 & 1.01 \\
    & & \cmark & \cmark &   & 70.39 & 83.92 & 88.50 & 92.63 & 95.94 & 1.01 & 71.17 & 84.03 & 88.85 & 92.77 & 95.76 & \textbf{1.00} \\
    & & \cmark &  & \cmark  & 74.95 & 87.52 & 91.42 & 94.71 & 97.22 & \textbf{1.00} & \textbf{75.71} & 87.66 & 91.70 & 94.98 & 97.42 & \textbf{1.00} \\
    & & \cmark &  &   & 74.45 & \textbf{87.91} & 92.36 & 95.35 & 97.79 & \textbf{1.00} & 75.00 & \textbf{88.09} & 92.38 & 95.85 & 98.13 & \textbf{1.00} \\
    \hline
    \end{tabular}}
    \vspace{-5pt}
\end{table}

\begin{table}[t]
\caption{Comparison of ``event $\rightarrow$ event'' retrieval results between the original TMR model and its variations equipped with different language models, which are fine-tuned with chronological negative samples.}
    \label{tab:fullres2}
    \vspace{-5pt}
    \centering
    \setlength{\tabcolsep}{4pt}
    \resizebox{0.99\linewidth}{!}{
    \begin{tabular}{llccc|cccccc|cccccc}
    \toprule \multirow{2}{*}{ Protocol } & \multirow{2}{*}{ Method }&\multirow{2}{*}{ Tune } &\multirow{2}{*}{ VAE }& \multirow{2}{*}{ Recon. }&  \multicolumn{6}{c}{ Text-to-motion retrieval } & \multicolumn{6}{|c}{ Motion-to-text retrieval } \\
     & & & & & $\mathrm{R} @ 1 \uparrow$ & $\mathrm{R} @ 2 \uparrow$ & $\mathrm{R} @ 3 \uparrow$ & $\mathrm{R} @ 5 \uparrow$ & $\mathrm{R} @ 10 \uparrow$ & MedR $\downarrow$ & $\mathrm{R} @ 1 \uparrow$ & $\mathrm{R} @ 2 \uparrow$ & $\mathrm{R} @ 3 \uparrow$ & R@5$\uparrow$ & $\mathrm{R} @ 10 \uparrow$ & MedR $\downarrow$ \\
    \hline\hline \multirow{17}{*}{ (a) All } 
    & TMR \cite{petrovich2023tmr} &  & \cmark & \cmark &5.54 & 10.49 & 14.01 & 20.39 & 32.30 & 24.00 & 9.08 & 11.52 & 16.77 & 22.51 & 33.99 & 23.50 \\ \cline{2-17}
    &  \multirow{4}{*}{ DistilBERT~\cite{sanh2019distilbert} }& \cmark & \cmark & \cmark   & 5.47 & 10.54 & 14.01 & 19.07 & 29.72 & 29.00 & 8.80 & 10.81 & 15.88 & 20.99 & 30.04 & 30.00 \\
    & & \cmark & \cmark &   & 4.24 & 8.96 & 12.04 & 17.88 &28.4 & 29.00 & 7.07 & 9.10 & 13.80 & 19.37 & 28.90 & 29.50 \\
    & & \cmark &  & \cmark  & 6.52 & 11.93 & 15.37 & 21.37 & 31.98 & 25.00 & 8.23 & 10.72 & 16.01 & 21.35 & 32.60 & 26.50 \\
    & & \cmark &  &   & 5.22 & 10.22 & 14.03 & 20.89 & 31.84 & 25.00 & 8.55 & 10.81 & 15.99 & 22.31 & 31.93 & 26.50 \\  \cline{2-17}
     &  \multirow{4}{*}{ CLIP \cite{radford2021learning} }& \cmark & \cmark & \cmark   & 4.95 & 9.51 & 12.77 & 18.50 & 28.70 & 30.00 & 8.12 & 10.08 & 14.71 & 20.05 & 29.93 & 30.00 \\
    & & \cmark & \cmark &   & 3.79 & 7.53 & 10.56 & 15.90 & 26.07 & 35.00 & 6.71 & 8.78 & 13.66 & 18.93 & 26.57 & 36.50 \\
    & & \cmark &  & \cmark  & 4.61 & 8.85 & 11.84 & 16.90 & 25.96 & 40.00 & 7.05 & 8.65 & 12.98 & 18.32 & 26.46 & 42.50 \\
    & & \cmark &  &   & 5.20 & 10.15 & 13.30 & 18.48 & 28.76 & 32.00 & 8.14 & 10.08 & 15.10 & 20.67 & 28.42 & 33.50 \\  \cline{2-17}
     &  \multirow{4}{*}{ t5-base \cite{raffel2020exploring} }& \cmark & \cmark & \cmark   & 5.00 & 10.31 & 13.73 & 19.14 & 29.77 & 29.00 & 7.92 & 10.10 & 14.69 & 20.51 & 30.04 & 29.50 \\
    & & \cmark & \cmark &   & 4.81 & 8.90 & 12.43 & 18.39 & 30.20 & 27.00 & 7.12 & 9.60 & 14.48 & 20.64 & 30.22 & 27.50 \\
    & & \cmark &  & \cmark  & 6.80 & 12.64 & 16.93 & 24.59 & 36.72 & 20.00 & 10.40 & 13.05 & 18.86 & 25.32 & 36.09 & 20.00 \\
    & & \cmark &  &   & 5.22 & 10.20 & 13.82 & 20.78 & 32.23 & 23.00 & 8.46 & 11.09 & 16.56 & 22.45 & 32.46 & 24.50 \\ \cline{2-17}
     & \multirow{4}{*}{ t5-large \cite{raffel2020exploring} }& \cmark & \cmark & \cmark   & 4.86 & 9.51 & 12.89 & 18.43 & 30.22 & 28.00 & 7.73 & 9.51 & 14.69 & 20.07 & 30.02 & 27.50 \\
    & & \cmark & \cmark &   & 4.70 & 8.87 & 12.20 & 18.64 & 30.20 & 26.00 & 7.37 & 9.60 & 14.83 & 20.80 & 31.09 & 26.00 \\
    & & \cmark &  & \cmark  & \textbf{7.21} & \textbf{14.26} & \textbf{18.39} & \textbf{25.34} & \textbf{38.55} & \textbf{19.00} & \textbf{10.93} & \textbf{13.89} & \textbf{19.80} & \textbf{26.44} & \textbf{37.34} & \textbf{20.00} \\
    & & \cmark &  &   & 5.86 & 11.41 & 15.31 & 22.63 & 35.10 & 21.00 & 8.90 & 11.18 & 17.22 & 23.45 & 35.24 & 21.50 \\
    
    \hline \multirow{17}{*}{ (b) All w/ threshold } 
    
    & TMR \cite{petrovich2023tmr} & & \cmark &  \cmark& 12.29 & 16.70 & 21.97 & 29.72 & 41.77 & 15.00 & 12.32 & 14.99 & 21.10 & 26.69 & 38.44 & 19.50 \\ \cline{2-17}
    &  \multirow{4}{*}{ DistilBERT~\cite{sanh2019distilbert} }& \cmark & \cmark & \cmark   & 11.86 & 14.99 & 20.78 & 27.26 & 37.77 & 20.00 & 12.43 & 14.26 & 20.23 & 26.03 & 35.36 & 23.50 \\
    & & \cmark & \cmark &   & 9.95 & 14.30 & 19.59 & 26.53 & 37.16 & 20.00 & 10.63 & 11.27 & 17.79 & 23.24 & 33.12 & 23.50 \\
    & & \cmark &  & \cmark  & 13.12 & 17.20 & 22.26 & 29.31 & 40.69 & 16.00 & 11.98 & 14.28 & 20.46 & 26.16 & 37.57 & 20.00 \\
    & & \cmark &  &  & 11.66 & 16.61 & 22.63 & 29.97 & 41.56 & 16.00 & 12.07 & 13.41 & 20.16 & 27.33 & 36.36 & 21.50 \\  \cline{2-17}
     &  \multirow{4}{*}{ CLIP \cite{radford2021learning} }& \cmark & \cmark & \cmark   & 10.99 & 14.69 & 20.46 & 27.01 & 38.12 & 20.00 & 11.41 & 13.44 & 19.07 & 24.34 & 34.58 & 23.50 \\
    & & \cmark & \cmark &   & 8.92 & 11.61 & 17.04 & 23.13 & 34.01 & 24.50 & 10.36 & 11.20 & 17.45 & 23.18 & 31.16 & 29.00 \\
    & & \cmark &  & \cmark  & 10.49 & 13.75 & 18.27 & 24.29 & 33.78 & 25.00 & 9.92 & 12.29 & 17.40 & 23.24 & 31.98 & 31.50 \\
    & & \cmark &  &   & 11.18 & 14.46 & 20.05 & 26.60 & 36.31 & 21.00 & 11.25 & 13.25 & 19.32 & 25.16 & 33.10 & 27.00 \\  \cline{2-17}
     &  \multirow{4}{*}{ t5-base \cite{raffel2020exploring} }& \cmark & \cmark & \cmark   & 11.91 & 15.81 & 21.33 & 28.03 & 39.23 & 18.00 & 12.07 & 13.16 & 19.46 & 25.21 & 34.67 & 23.25 \\
    & & \cmark & \cmark &   & 11.84 & 16.24 & 22.17 & 29.11 & 42.13 & 18.00 & 10.45 & 11.36 & 17.15 & 23.70 & 33.60 & 23.50 \\
    & & \cmark &  & \cmark  & 13.71 & 19.30 & 25.94 & 34.99 & 46.85 & \textbf{12.00} & 14.32 & 16.79 & 23.52 & 30.25 & 40.26 & 17.00 \\
    & & \cmark &  &   & 11.82 & 16.33 & 21.85 & 30.20 & 41.86 & 15.00 & 11.68 & 13.73 & 20.05 & 26.37 & 36.66 & 20.00 \\ \cline{2-17}
     & \multirow{4}{*}{ t5-large \cite{raffel2020exploring} }& \cmark & \cmark & \cmark   & 11.59 & 15.40 & 20.83 & 27.46 & 40.72 & 17.00 & 11.61 & 12.16 & 18.96 & 24.75 & 34.19 & 23.00 \\
    & & \cmark & \cmark &   & 11.77 & 15.49 & 20.73 & 28.74 & 40.33 & 16.00 & 10.86 & 12.02 & 18.75 & 25.14 & 35.15 & 22.00  \\
    & & \cmark &  & \cmark  & \textbf{14.23} & \textbf{20.37} & \textbf{26.30} & \textbf{34.33} & \textbf{47.90} & \textbf{12.00} & \textbf{14.71} & \textbf{17.29} & \textbf{24.68} & \textbf{31.36} & \textbf{41.67} & \textbf{16.00} \\
    & & \cmark &  &   & 12.86 & 18.09 & 23.63 & 32.39 & 45.99 & 13.00 & 11.91 & 13.66 & 20.94 & 27.62 & 39.37 & 17.50 \\

    \hline \multirow{17}{*}{ (c) Dissimilar subset } 
    & TMR \cite{petrovich2023tmr} & & \cmark & \cmark & 52.00 & 66.00 & 74.00 & 81.00 & 88.00 & 1.00 & 46.00 & 68.00 & 75.00 & 81.00 & 86.00 & 2.00 \\ \cline{2-17}
    &  \multirow{4}{*}{ DistilBERT~\cite{sanh2019distilbert} }& \cmark & \cmark & \cmark   & 50.00 & 73.00 & 78.00 & 85.00 & 90.00 & 1.50 & 52.00 & 69.00 & 75.00 & 82.00 & 89.00 & \textbf{1.00} \\
    & & \cmark & \cmark &   & 47.00 & 66.00 & 74.00 & 77.00 & 85.00 & 2.00 & 48.00 & 69.00 & 75.00 & 85.00 & 85.00 & 2.00 \\
    & & \cmark &  & \cmark  & 53.00 & 67.00 & 76.00 & 85.00 & 93.00 & \textbf{1.00} & 52.00 & 68.00 & \textbf{80.00} & 85.00 & 91.00 & \textbf{1.00} \\
    & & \cmark &  &   & 48.00 & 68.00 & 78.00 & 85.00 & 90.00 & 2.00 & 47.00 & 68.00 & 79.00 & 86.00 & 90.00 & 2.00 \\  \cline{2-17}
     &  \multirow{4}{*}{ CLIP \cite{radford2021learning} }& \cmark & \cmark & \cmark  & 46.00 & 64.00 & 72.00 & 85.00 & 91.00 & 2.00 & 44.00 & 69.00 & 74.00 & 84.00 & 90.00 & 2.00 \\
    & & \cmark & \cmark &   & 50.00 & 65.00 & 73.00 & 75.00 & 86.00 & 1.50 & 49.00 & 68.00 & 71.00 & 77.00 & 83.00 & 2.00 \\
    & & \cmark &  & \cmark  & 44.00 & 61.00 & 69.00 & 76.00 & 80.00 & 2.00 & 45.00 & 58.00 & 66.00 & 75.00 & 80.00 & 2.00 \\
    & & \cmark &  &   & 47.00 & 62.00 & 73.00 & 79.00 & 84.00 & 2.00 & 46.00 & 62.00 & 72.00 & 76.00 & 86.00 & 2.00 \\  \cline{2-17}
     &  \multirow{4}{*}{ t5-base \cite{raffel2020exploring} }& \cmark & \cmark & \cmark   & 49.00 & 67.00 & 74.00 & 82.00 & 89.00 & 2.00 & 44.00 & 62.00 & 73.00 & 82.00 & 89.00 & 2.00 \\
    & & \cmark & \cmark &   & 53.00 & 67.00 & 76.00 & 85.00 & \textbf{95.00} & \textbf{1.00} & 53.00 & 67.00 & 75.00 & 86.00 & 94.00 & \textbf{1.00} \\
    & & \cmark &  & \cmark  & 49.00 & 71.00 & 77.00 & \textbf{87.00} & 94.00 & 2.00 & 51.00 & 69.00 & 79.00 & \textbf{88.00} & \textbf{95.00} & \textbf{1.00} \\
    & & \cmark &  &   & 51.00 & 67.00 & 74.00 & 83.00 & 94.00 & \textbf{1.00} & 50.00 & 66.00 & 75.00 & 83.00 & 92.00 & 1.50 \\ \cline{2-17}
     & \multirow{4}{*}{ t5-large \cite{raffel2020exploring} }& \cmark & \cmark & \cmark   & 45.00 & 67.00 & 76.00 & 86.00 & 92.00 & 2.00 & 50.00 & \textbf{73.00} & \textbf{80.00} & 86.00 & 92.00 & 1.50 \\
    & & \cmark & \cmark &   & 48.00 & 66.00 & 70.00 & 82.00 & 94.00 & 2.00 & 54.00 & 66.00 & 73.00 & 85.00 & 91.00 & \textbf{1.00} \\
    & & \cmark &  & \cmark  & \textbf{59.00} & \textbf{77.00} & \textbf{80.00} & \textbf{87.00} & 94.00 & \textbf{1.00} & 53.00 & \textbf{73.00} & 79.00 & 86.00 & 93.00 & \textbf{1.00} \\
    & & \cmark &  &   & 54.00 & 72.00 & 77.00 & \textbf{87.00} & \textbf{95.00} & \textbf{1.00} & \textbf{58.00} & 69.00 & 76.00 & 86.00 & 94.00 & \textbf{1.00}  \\

    \hline \multirow{17}{*}{ (d) Small batches } 
    & TMR \cite{petrovich2023tmr} & &\cmark &\cmark& 69.37 & 84.17 & 89.58 & 93.45 & 97.06 & 1.00 & 70.00 & 84.90 & 90.01 & 93.68 & 97.03 & 1.02 \\ \cline{2-17}
    &  \multirow{4}{*}{ DistilBERT~\cite{sanh2019distilbert} }& \cmark & \cmark & \cmark   & 64.83 & 78.85 & 84.19 & 89.10 & 94.02 & 1.05 & 64.21 & 77.71 & 82.80 & 87.93 & 92.52 & 1.05 \\
    & & \cmark & \cmark &   & 66.70 & 80.59 & 85.99 & 89.90 & 93.34 & 1.03 & 66.42 & 79.74 & 85.06 & 89.03 & 92.79 & 1.01 \\
    & & \cmark &  & \cmark  & 67.75 & 81.55 & 86.43 & 90.81 & 95.48 & 1.01 & 67.47 & 80.93 & 85.52 & 90.05 & 94.82 & 1.03 \\
    & & \cmark &  &   & 69.39 & 84.10 & 88.64 & 92.20 & 95.96 & 1.00 & 68.32 & 82.98 & 88.05 & 91.65 & 95.62 & 1.01  \\  \cline{2-17}
     &  \multirow{4}{*}{ CLIP \cite{radford2021learning} }& \cmark & \cmark & \cmark   & 65.90 & 80.70 & 86.02 & 91.17 & 95.10 & 1.03 & 66.42 & 80.45 & 86.09 & 90.42 & 94.91 & 1.02 \\
    & & \cmark & \cmark &   & 62.73 & 75.96 & 81.66 & 86.11 & 91.24 & 1.08 & 62.00 & 75.16 & 80.27 & 84.99 & 89.37 & 1.08 \\
    & & \cmark &  & \cmark & 59.33 & 74.32 & 80.09 & 86.41 & 92.95 & 1.14 & 58.96 & 72.65 & 78.92 & 84.06 & 91.01 & 1.17 \\
    & & \cmark &  &   & 64.03 & 77.24 & 82.16 & 87.07 & 92.75 & 1.05 & 63.23 & 75.80 & 80.57 & 85.61 & 91.38 & 1.07 \\  \cline{2-17}
     &  \multirow{4}{*}{ t5-base \cite{raffel2020exploring} }& \cmark & \cmark & \cmark   & 68.02 & 83.26 & 88.73 & 93.39 & 96.78 & 1.02 & 68.61 & 83.49 & 89.14 & 93.48 & 96.90 & 1.01 \\
    & & \cmark & \cmark &   & 69.02 & 83.65 & 89.67 & 93.29 & 96.74 & 1.01 & 69.30 & 83.62 & 89.55 & 93.59 & 96.76 & \textbf{1.00} \\
    & & \cmark &  & \cmark  & 73.29 & \textbf{86.93} & 91.56 & 95.07 & \textbf{97.97} & 1.01 & 72.22 & \textbf{86.54} & \textbf{91.35} & 94.53 & 97.24 & \textbf{1.00} \\
    & & \cmark &  &   & 70.53 & 84.42 & 89.85 & 94.07 & 97.15 & 1.01 & 70.14 & 83.94 & 89.42 & 92.93 & 95.92 & 1.02 \\ \cline{2-17}
     & \multirow{4}{*}{ t5-large \cite{raffel2020exploring} }& \cmark & \cmark & \cmark   & 69.05 & 83.74 & 89.74 & 94.23 & 97.29 & 1.02 & 69.71 & 84.31 & 89.96 & 94.25 & 97.15 & 1.01 \\
    & & \cmark & \cmark &   & 70.39 & 83.92 & 88.50 & 92.63 & 95.94 & 1.01 & 71.17 & 84.03 & 88.85 & 92.77 & 95.76 & \textbf{1.00} \\
    & & \cmark &  & \cmark  & \textbf{74.00} & 86.63 & 90.67 & 94.23 & 97.15 & \textbf{1.00} & \textbf{73.31} & 85.74 & 90.42 & 93.82 & 96.58 & \textbf{1.00}  \\
    & & \cmark &  &   &  73.45 & \textbf{86.93} & \textbf{91.77} & \textbf{95.23} & 97.92 & \textbf{1.00} & 72.72 & 86.52 & 91.17 & \textbf{95.07} & \textbf{97.56} & 1.01 \\
    \hline
    \end{tabular}}
    \vspace{-5pt}
\end{table}
\end{document}